\documentclass[10pt,twocolumn,letterpaper]{article}

\usepackage[pagenumbers]{wacv} 

\usepackage{graphicx}
\usepackage{dsfont}
\usepackage{amsmath,amssymb,amstext,amsfonts, bm, dsfont}
\usepackage{amssymb}
\usepackage{booktabs}
\usepackage{multirow}
\usepackage[table]{xcolor}
\usepackage{float}
\usepackage{tabularx}
\usepackage[accsupp]{axessibility} 
\usepackage[pagebackref,breaklinks,colorlinks]{hyperref}

\usepackage[capitalize]{cleveref}
\crefname{section}{Sec.}{Secs.}
\Crefname{section}{Section}{Sections}
\Crefname{table}{Table}{Tables}
\crefname{table}{Tab.}{Tabs.}

\def\wacvPaperID{702} 
\def\confName{WACV}
\def\confYear{2025}

\begin{document}

\title{DrIFT: Autonomous \underline{Dr}one Dataset with \underline{I}ntegrated Real and Synthetic Data, \underline{F}lexible Views, and \underline{T}ransformed Domains}

\author{Fardad Dadboud$^{1}$, Hamid Azad$^{1}$, Varun Mehta$^2$, Miodrag Bolic$^1$, Iraj Mantegh$^2$\\
$^1$University of Ottawa, $^2$National Research Council Canada\\
}
\maketitle

\begin{figure*}
    \begin{subfigure}{0.49\linewidth}
        \centering
        \includegraphics[width=\linewidth]{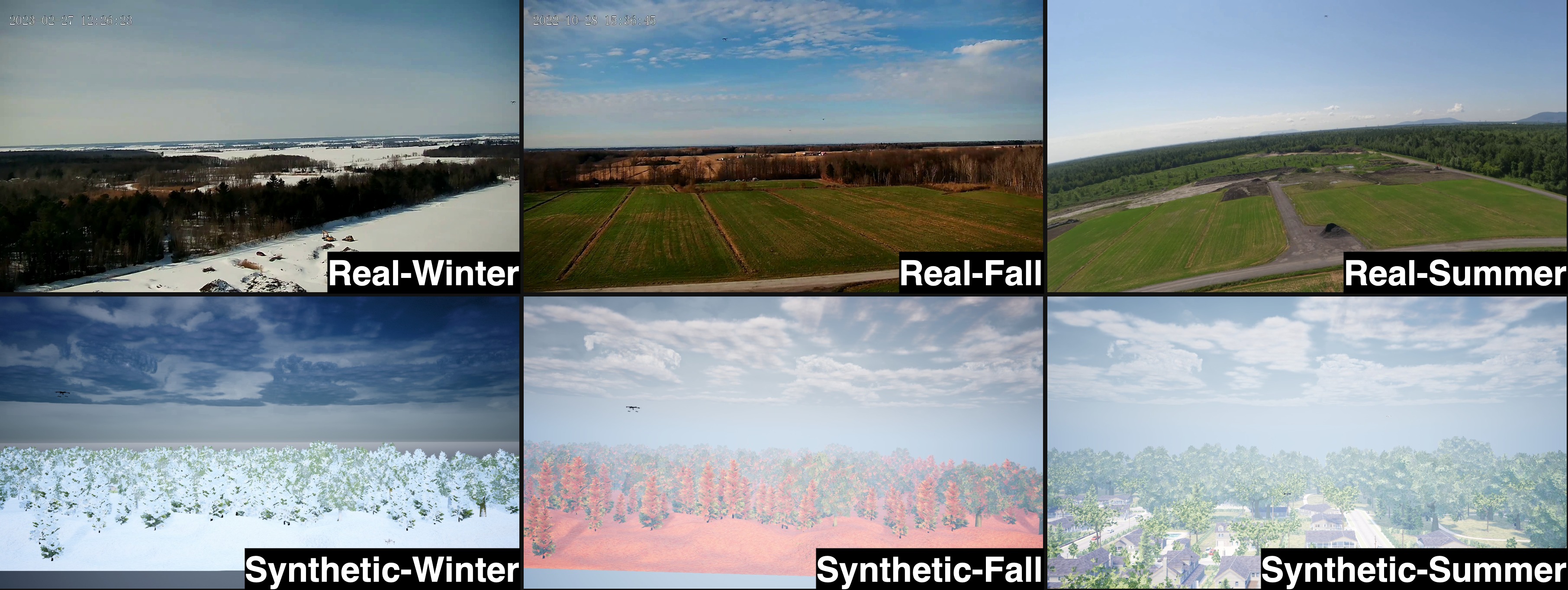}
        \caption{Aerial}
        \label{fig:samp_aerial}
    \end{subfigure}
    \hfill
    \begin{subfigure}{0.49\linewidth}
        \centering
        \includegraphics[width=\linewidth]{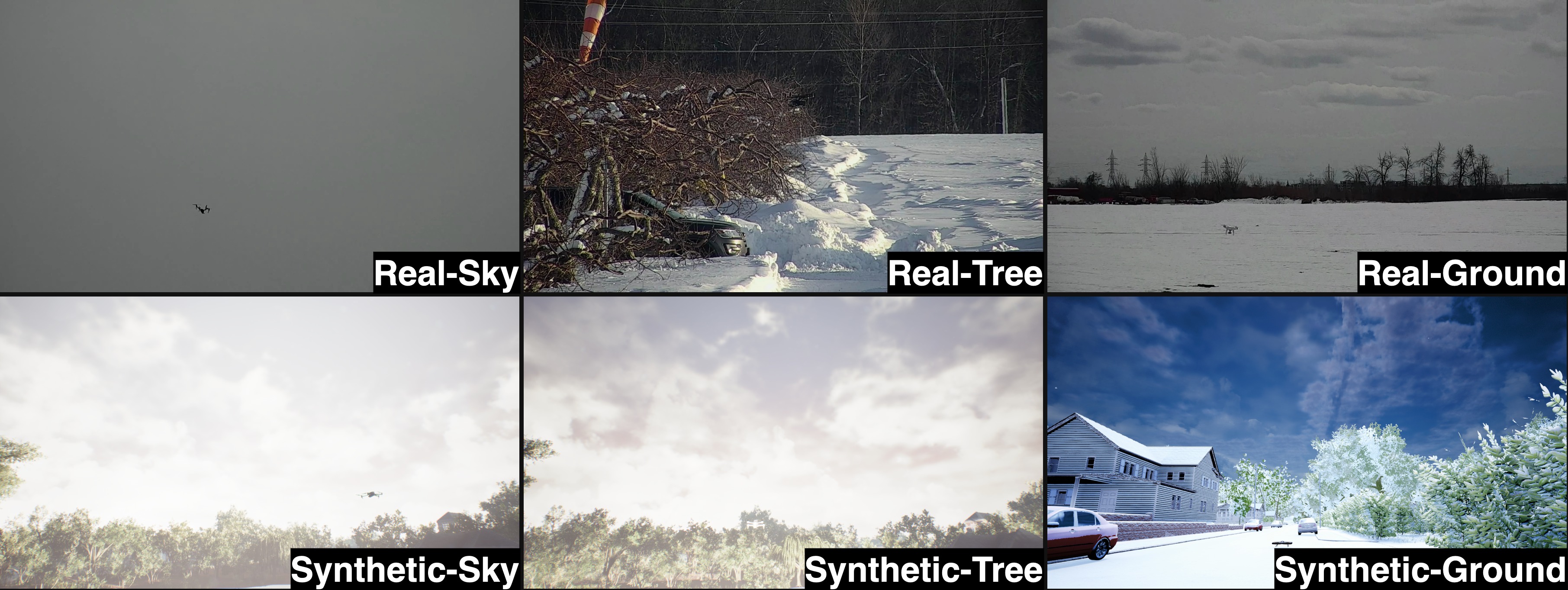}
        \caption{Ground(Winter)}
        \label{fig:samp_ground}
    \end{subfigure}
    \begin{subfigure}{0.49\linewidth}
        \centering
        \includegraphics[width=\linewidth]{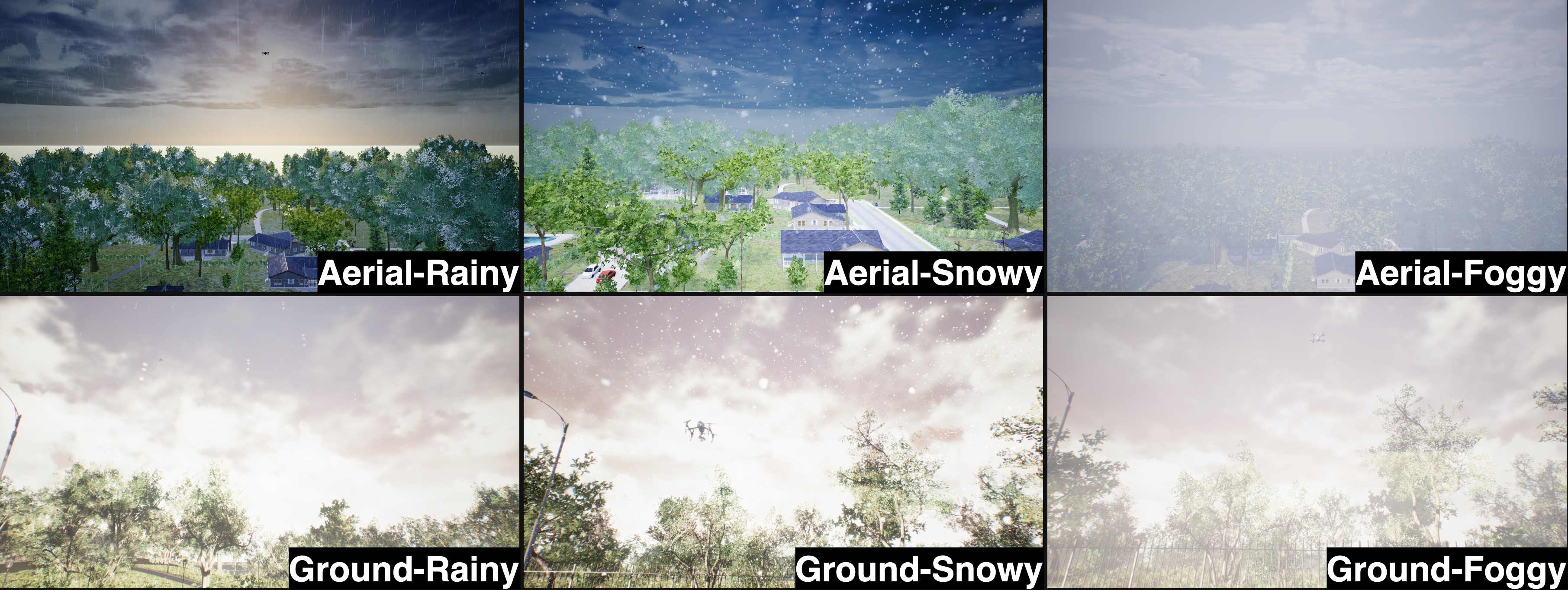}
        \caption{Adverse(Synthetic)}
        \label{fig:samp_adv}
    \end{subfigure}
    \hfill
    \begin{subfigure}{0.49\linewidth}
        \centering
        \includegraphics[width=\linewidth]{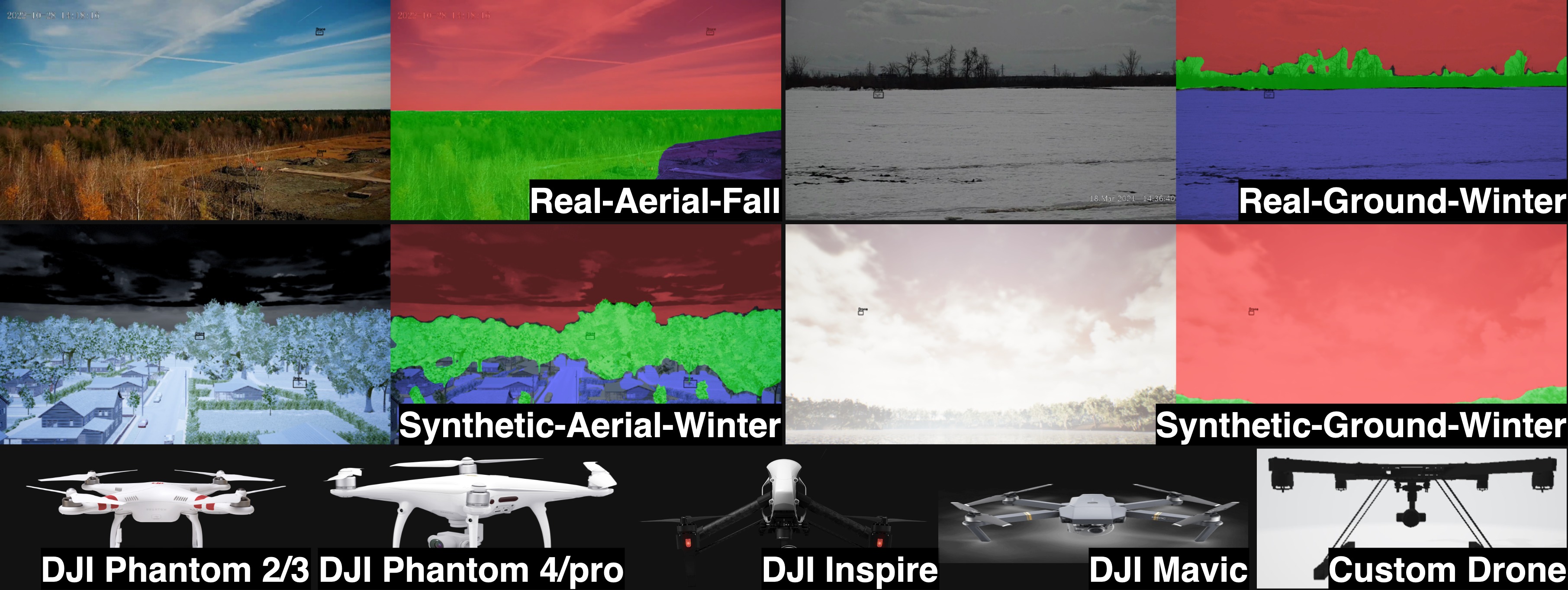}
        \caption{Ground Truth and Drone Models}
        \label{fig:samp_gt_drones}
    \end{subfigure}
    \caption{Samples of the DrIFT dataset: a) aerial PoVs in different seasons for both real and synthetic data, b) ground PoV real and synthetic data that has been recorded in winter with a sky, tree, or ground background, c) adverse weather in a synthetic environment for both aerial and ground PoV data, and d) ground truth bounding boxes and background segmentation maps have been illustrated in first two rows. The utilized drones have been also depicted in the last row.}
    \label{fig:samples}
\end{figure*}
\begin{abstract}
    Dependable visual drone detection is crucial for the secure integration of drones into the airspace. However, drone detection accuracy is significantly affected by domain shifts due to environmental changes, varied points of view, and background shifts. To address these challenges, we present the DrIFT dataset, specifically developed for visual drone detection under domain shifts. DrIFT includes fourteen distinct domains, each characterized by shifts in point of view, synthetic-to-real data, season, and adverse weather. DrIFT uniquely emphasizes background shift by providing background segmentation maps to enable background-wise metrics and evaluation. Our new uncertainty estimation metric, MCDO-map, features lower postprocessing complexity, surpassing traditional methods. We use the MCDO-map in our uncertainty-aware unsupervised domain adaptation method, demonstrating superior performance to SOTA unsupervised domain adaptation techniques. The dataset is available at: \href{https://github.com/CARG-uOttawa/DrIFT.git}{https://github.com/CARG-uOttawa/DrIFT.git}.\hfill
\end{abstract}

\section{Introduction}
\label{sec:intro}
Uncrewed Aerial Vehicles (UAVs), also known as drones, have gained popularity in recent years due to their versatility and cost-effectiveness for various operations\cite{floreano2015science,zhang2012application}, including healthcare \cite{restas2022drone}, surveillance \cite{dilshad2020applications, choi2020unsupervised}, delivery \cite{MOSHREFJAVADI2021114854}, agriculture \cite{budiharto2019review, albert2022unsupervised}, construction and mining  \cite{shahmoradi2020comprehensive, drones7080515}, infrastructure inspection \cite{al2017vbii}, and search-and-rescue \cite{MOHDDAUD202230}. However, their ubiquitous use has raised safety concerns, such as the possibility of their use for malicious activities and collisions with other objects in the airspace \cite{khan2022detection,taha2019machine}. Achieving autonomous flight capabilities in challenging environments,  for individual and swarm drones, is vital for various applications \cite{huang2017structure}.
Ensuring the safety of such operations depends on the accurate and efficient processing of drone-related data. 

In particular, vision-based drone detection plays a crucial role, as it faces challenges such as distant small objects, handling complex backgrounds (BGs), and distinguishing drones from other visually similar flying objects. Deep Neural Networks (DNNs) have demonstrated exceptional capabilities in multiple applications, including drone detection \cite{chen2017deep, coluccia2021drone, delleji2022upgraded, jiang2021antiuav}. However, distribution shifts from the training to the test set, caused by environmental variations, various points of view (PoVs), and background changes, pose intrinsic challenges in drone detection and affect the DNNs capabilities. Specifically, BG shift, \eg training with data mostly captured with sky background while sky, tree, and ground backgrounds appear in the validation set, is also called unseen BG \cite{zhang2021meta}. Gathering supervised data for all domains to ensure DNN generalization is impractical and often costly for data collection and annotation \cite{litrico2023guiding, roy2022uncertainty, munir2021ssal}, especially, it is worsened in adverse weather conditions or under regulatory constraints for drone-based applications.

Unsupervised Domain Adaptation (UDA) \cite{munir2021ssal, chen2022learning, chen2018domain, guan2021uncertainty, nguyen2020domain} is a principal approach to addressing domain shift (DS) in object detection (OD). Domain shift refers to shifts in the input image due to environmental factors that affect the performance of drone detection due to their impact on the drone's appearance in the scene. UDA aims to transfer knowledge from the source to the target domain, despite the lack of supervision in the target domain. This approach has gained popularity in applications such as autonomous vehicles \cite{sun2022shift, Franchi2022MUAD, SDV21, SDHV18} and other edge-AI, where DSs are common, and supervised data is not guaranteed \cite{murphy2022probabilistic, csurka2022visual}. UDA methods have also been employed extensively to address DSs in drone detection \cite{rozatsev2019beyond, rui2021comprehensive}. However, unforeseen situations that cause DSs, such as drones with novel shapes, can still occur. Despite the trend toward using UDA in the field, there is a lack of comprehensive exploration of specialized domain shift and UDA methodologies in drone detection \cite{rozatsev2019beyond, sun2020tib, jiang2021antiuav, zheng2021air, rozantsev2015flying, chen2017deep, walter2020training, svanstrom2021real, rodriguez2020adaptive, zhao2022vision}. This gap has catalyzed our work to design a new dataset that addresses these specific challenges.

Combining existing datasets often results in multiple uncontrolled DSs co-occurring, making it difficult to isolate and examine the impact of specific shifts. Moreover, existing datasets lack systematic background segmentation and comprehensive coverage of DS types, making manual annotations costly and infeasible. To overcome this, the DrIFT dataset was designed to provide a controlled environment in which individual DS types, PoV, season, weather, and background can be studied independently. DrIFT ensures a balanced distribution across its fourteen distinct domains, addressing limitations in previous datasets and enabling the systematic study of multiple DS types simultaneously. Driven by the need to address these shortages, we present the DrIFT dataset with the following pivotal contributions:

\begin{enumerate}
    \item The DrIFT dataset introduces a vision-based drone detection dataset. Uniquely, DrIFT comprises fourteen distinct domains constructed by combinations of four major domain shifts: PoV, synthetic-to-real, season, and weather. In most domains, there are sky, trees, and ground backgrounds.
    \item We employ BG segmentation maps to introduce the concept of BG shift as a distinct challenge. This novel approach allows us to report BG-wise metrics (\eg, $AP_{50}^{S}$:$AP_{50}$ of sky background detections), providing a focused study on how 
    BG shift
    influence the object detection.
    \item We introduce a novel uncertainty evaluation method for OD, surpassing existing methods (\cref{tab:unccomp}). Our method, utilizing a score map, offers significant advantages such as lower complexity of postprocessing and superior capability in capturing DS.
    \item Our uncertainty-aware UDA method 
    outperforms state-of-the-art (SOTA) UDA methods for drone detection (\cref{tab:udares}).
\end{enumerate}

\begin{table}
    \centering
    {\small{
    \begin{tabular}{l|c|c|c|c}
        \toprule 
        Dataset & Target & Real/Synt. & PoV & \#DS\\
        \midrule
        \cite{coluccia2021drone} & Multiple & Real & Gr. & $0$\\
        \cite{sun2020tib} & Single & Real & Gr. & $0$ \\
        \cite{jiang2021antiuav} & Single & Real & Gr. & $0$\\
        \cite{zheng2021air} & Single & Real & Aerial & $0$ \\
        \cite{rozantsev2015flying} & Multiple & Real & Aerial & $0$ \\
        \cite{rozatsev2019beyond} & Multiple & Real+Synt. & - & $1$ \\
        \cite{chen2017deep} & Multiple & Real & Gr. & $0$ \\
        \cite{walter2020training} & Multiple & Real & Aerial & $0$ \\
        \cite{svanstrom2021real} & Single & Real & Gr. & $0$ \\
        \cite{rodriguez2020adaptive} & Single & Real & Gr. & $0$ \\
        \cite{zhao2022vision} & Multiple & Real & Gr. & $0$ \\
        \cite{ajakwe2022visiodect} & Multiple & Real & Gr. & $0$ \\
        \cite{barisic2022sim2air} & Multiple & Synt. & Gr.+Aerial & $0$ \\
        \cite{delleji2022upgraded} & Single & Real & Gr. & $0$ \\
        \textbf{DrIFT} & \textbf{Multiple} & \textbf{Real+Synt.} & \textbf{Gr.+Aerial} & $\mathbf{4}$ \\
        \bottomrule
    \end{tabular}
    }}
    \caption{Drone Datasets: The DrIFT dataset, as the first drone detection dataset to study four DSs, includes image frames with multiple drones, real and synthetic data, and ground and aerial PoVs. The "\#DS" indicates the number of types of DSs studied.}
    \label{tab:datasets}
\end{table}

\section{Related Work}
\label{sec:RelWork}

\subsection{Drone Datasets}
Drone datasets have recently become publicly available to address the increasing interest in drone detection \cite{jiang2021review}. Many datasets have significant limitations, as highlighted in \cref{tab:datasets}. For instance, the dataset in \cite{coluccia2021drone} lacks certain weather conditions and uses a stationary camera. The datasets in \cite{sun2020tib} and \cite{jiang2021antiuav} are limited to a single PoV. The dataset in \cite{zheng2021air} is restricted to partly cloudy and clear weather. Other datasets \cite{rozantsev2015flying, chen2017deep, walter2020training} offer limited diversity in weather and PoV. UAV-200 \cite{rozatsev2019beyond} uses supervised domain adaptation with a fraction of the target domain during training and only examines the synthetic-to-real DS, while DrIFT studies four types of domain shift in a UDA manner. The number of DS types studied is indicated in \cref{tab:datasets}. The datasets in \cite{svanstrom2021real} and \cite{zhao2022vision} feature multiple drone models but lack comprehensive DSs. The datasets in \cite{ajakwe2022visiodect} and \cite{delleji2022upgraded} focus primarily on ground PoV videos with limited weather conditions. \cite{barisic2022sim2air} lacks real-world domain. 

The \textbf{DrIFT dataset} introduces fourteen distinct domains constructed by combinations of four major DS elements: PoV, synthetic-to-real, season, and weather, with sky, trees, and ground backgrounds (\cref{fig:samples}). DrIFT uniquely emphasizes BG shift as a separate challenge and employs BG segmentation maps to create BG-wise metrics. This comprehensive approach addresses the lack of datasets that study various DSs in drone detection, making DrIFT the first dataset to comprehensively study all four DSs.

\subsection{Land Vehicle Datasets}
Land vehicle datasets constitute another topic similar to those of drones within the realm of autonomous vehicles. As inspiration for DrIFT, the SHIFT \cite{sun2022shift} autonomous driving dataset offers DSs across a spectrum of parameters, such as weather conditions, time of day, and density of vehicle and pedestrian, but does not investigate BG shifts.

\subsection{Uncertainty Estimation}
Uncertainty estimation is crucial for assessing the safety level of autonomous vehicles, especially drones, by effectively dealing with DS. Conventional methods categorize uncertainty in deep learning into aleatoric and epistemic uncertainties. Aleatoric uncertainty arises from data noise, while epistemic uncertainty is due to limited data or domain coverage, which is more relevant to DS \cite{feng2021review}.

Historically, uncertainty estimation involves sampling-based techniques like Monte Carlo dropout (MCDO) \cite{gal2016dropout}, which, although effective in capturing epistemic uncertainty, are computationally intensive due to their iterative nature and postprocessing complexity \cite{harakeh2020bayesod}. 

To address computational constraints and accurately capture the epistemic uncertainty arising from data gaps, recent studies \cite{riedlinger2022uncertainty, riedlinger2023gradient} explore using gradient self-information directly to assess uncertainty. Nevertheless, they do not inherently encompass the true essence of uncertainty.

To address these computational constraints and the lack of a comprehensive sense of uncertainty, we leverage an efficient approach that combines the strengths of MCDO with a simplified postprocessing mechanism. Our method utilizes MCDO to generate uncertainty maps for each detection, performing multiple inference passes and aggregating these uncertainties into an overall score map \cite{oksuz2023towards}, which reduces the postprocessing complexity (\cref{subsec:meth}).

\subsection{Detection Calibration Error Estimation (D-ECE)}
D-ECE is critical for providing accurate confidence assessments in neural networks, especially for safety-critical applications. The calibration error measures the alignment between the predicted confidence and the actual results, helping to assess the reliability of a model \cite{kuppers2020multivariate, harakeh2021estimating}. D-ECE extends from classification-based calibration error estimation but applies specifically to detection tasks, focusing on the regression outputs of object detectors. The concept, introduced by \cite{kuppers2020multivariate}, addresses unique detection confidence calibration errors. Further details on its calculation are provided in \cref{subsec:meth}.


\subsection{Unsupervised Domain Adaptation (UDA)}
UDA addresses domain shift by transferring knowledge from a labeled source domain to an unlabeled target domain. UDA for object detection was first introduced by \cite{chen2018domain}.

Many approaches in UDA have been introduced that come with notable limitations. Pseudo-labeling and self-training methods, such as \cite{kim2019self} and \cite{khodabandeh2019robust}, generate target pseudo-labels, but incorrect labels can propagate errors, especially in complex backgrounds like our application. Image-to-image translation techniques \cite{hoffman2018cycada, sankaranarayanan2018generate} reduce the domain gap by converting source images into the target style, but these often introduce artifacts and require extensive training data to perform well, which is not feasible in our application.

Among the more recent advancements, uncertainty-aware methods have gained attention for their ability to improve domain adaptation by estimating and incorporating prediction uncertainties. These methods, such as \cite{munir2021ssal,guan2021uncertainty,nguyen2020domain, chen2022learning}, leverage uncertainty metrics to focus on areas where domain shifts are most pronounced. Adversarial training, introduced by \cite{chen2018domain}, complements this by aligning feature distributions between domains. Together, these approaches provide a robust mechanism for handling domain shift, focusing on confident regions and learning domain-invariant features to reduce errors and enhance model robustness. The details of our approach are discussed further in \cref{subsec:meth}.

\section{DrIFT Dataset}
\label{sec:drift}
We have developed a vision-based drone detection dataset consisting of image frames, ground truth bounding boxes, and BG segmentation maps (\cref{fig:samp_gt_drones}). In \cref{subsec:overview}, an overview of the DrIFT dataset's sensor, experimental setup, annotation, and dataset design has been presented. In the following, precise information regarding DrIFT's various domains has been compiled to represent the dataset's purpose for the DS. For more detailed statistics of DrIFT, the reader can go through the supplementary materials.
\subsection{The DrIFT Story}
\label{subsec:overview}
\textbf{Real Ground PoV's} video recordings for the DrIFT dataset were captured with a Bosch pan-tilt-zoom (PTZ) camera. The DJI Phantom 2/3, Phantom 4/Pro, Inspire, and Mavic (\cref{fig:samp_gt_drones}) were captured between 0.1 and 1.5 kilometers away in the recordings. A drone is predominantly present in the frames. The semi-automatic annotation has been done using the CVAT \cite{CVAT_ai_Corporation_Computer_Vision_Annotation_2022}. We have generated multiple other domains of data in our dataset to represent the DS.

\textbf{Real Aerial PoV} has been added to the DrIFT dataset to achieve the PoV shift concept. For the aerial PoV, a custom-built drone model was utilized (\cref{fig:samp_gt_drones}). 
In this experiment, mobile electro-optical cameras, the Infiniti STR-8MP-3X and GoPro were used to record multiple drone footage between 20 and 100 meters in the line of sight. The frames were recorded in different seasons, resulting in various BGs, such as the sky, trees in various seasons, and the ground with different colors.

\textbf{Synthetic Data} is recorded in the AirSim \cite{shah2018airsim} simulator for simulating real-world data counterparts in a simulated environment for all domains for considering synthetic-to-real domain shift, and due to the impossibility of flying in adverse weather conditions.

It is a common practice for domain-adaptive network training to have the same number of samples in the source and target domains \cite{Cordts2016Cityscapes}. 
Therefore, we designed the DrIFT dataset to maintain a balanced number of samples across domains within both the training and validation sets as long as we had sufficient real data for the domains.

\textbf{Background segmentation,} as one of the contributions of the DrIFT dataset, is important for its innovative exploration of BG shift. All validation frames' backgrounds have been segmented into sky, tree, and ground segments (\cref{fig:samp_gt_drones}) using the Track Anything platform \cite{yang2023track,kirillov2023segment}. By utilizing segmentation maps, it becomes feasible to utilize different metrics corresponding to different backgrounds (\cref{tab:shiftresults}, details in \cref{subsec:meth}).

All annotations were then double-checked and refined by human annotators to ensure accuracy.




\subsection{Dataset Design}
\label{subsec:datasetdesign}
To address a deficiency in drone detection datasets, we designed DrIFT with a concentration on studying common domain shifts in the wild. 

\textbf{Synthetic-to-Real:} In practical scenarios, capturing every conceivable real-world situation can be infeasible due to logistical challenges, resource limitations, and the prohibitive costs of annotation. To this end, we brought up synthetic data in order to initiate research on synthetic-to-real DS.
In DrIFT, all real-world data domains have simulated counterparts except for adverse weather conditions that do not exist in our real-world part of the dataset.

\textbf{PoV Shift:} The camera's PoV change (ground and aerial) contains different BGs and orientations of the target objects. This shift can significantly impact detection performance, making it a distinct type of DS.

\textbf{Weather Shift:} drones cannot be easily deployed in adverse weather. On the other hand, because this is a common DS in the wild, the system must be robust. Therefore, synthetic data is collected to study weather DS. 

\textbf{Background Shift:} The unseen background problem \cite{zhang2021meta}, also called BG shift in DrIFT, is present in various drone detection or autonomous driving datasets regarding the aforementioned DS. Nevertheless, no study has explicitly looked into the BG shift in object detection using BG segmentation maps. DrIFT investigates the BG shift from the sky to the tree and ground.

\section{DrIFT Benchmark}
\label{sec:bench}
This section first provides a comprehensive overview of the methodology used for the benchmark.
The following subsection provides a comprehensive overview of the different benchmark scenarios. Subsequently, the results of the benchmark are reported.
This section concludes with a comprehensive analysis of the benchmark outcomes and the dominant challenges of the DrIFT dataset. The supporting statements will be presented in the supplementary materials.

\subsection{Methodology}
\label{subsec:meth}
\begin{figure*}[!ht]
    \centering
    \includegraphics[width=\linewidth]{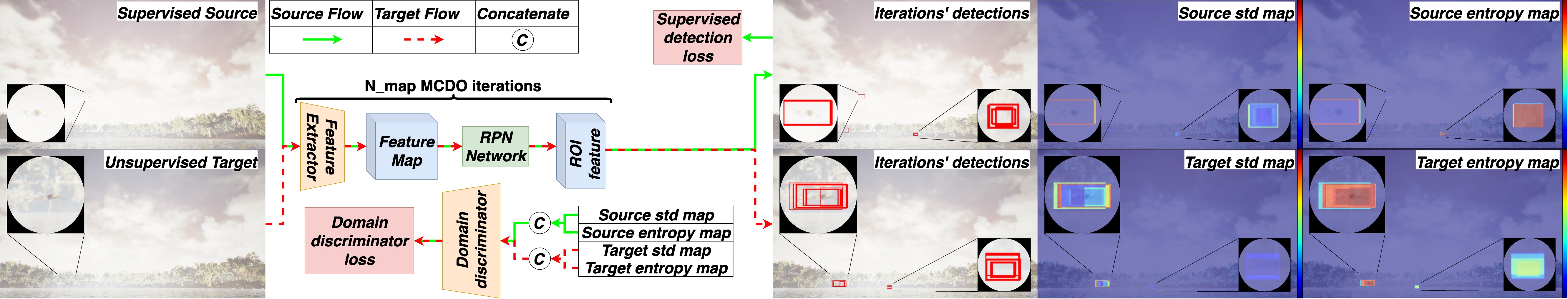}
    \caption{Uncertainty-aware UDA framework: In addition to supervised learning on the source domain, the concatenated std and entropy maps are input into the discriminator as part of the adversarial learning process. Magnified regions around detections are shown for better visualization. Colorbars are placed on the right side of the std and entropy maps. Green-solid and red-dash lines represent the source and target domain paths, respectively. All detections from multiple iterations are displayed to illustrate the generation and behavior of the uncertainty maps.}
    \label{fig:uda}
\end{figure*}
\subsubsection{Problem definition}
The primary goal of the DrIFT benchmark is to evaluate the performance of OD models under various shifts and the capabilities of UDA methods to address this issue.
Performance metrics include average precision (AP), uncertainty metrics, and D-ECE which are reported BG-wise.

Let $ \mathcal{D} = \{(\bm{X}_{\mathrm{i}}, \mathds{Y}_{\mathrm{i}})\}_{\mathrm{i}=1}^{\mathrm{N}} $ be the dataset, where $\bm{X}_{\mathrm{i}}$ are the input images and $\mathds{Y}_{\mathrm{i}}$ are the set of ground truth annotations containing bounding box coordinates and class labels for objects within each input image. The OD model predicts a set of detections $ \hat{\mathds{Y}}_{\mathrm{i}} = \{\hat{\bm{y}}^\mathrm{j}_\mathrm{i}\}_{\mathrm{j}=1}^{\mathrm{N}_{x_{\mathrm{i}}}} $, where each detection $ \hat{\bm{y}}^\mathrm{j}_\mathrm{i} = (\hat{\bm{b}}^\mathrm{j}_\mathrm{i}, \hat{c}^\mathrm{j}_\mathrm{i}, \hat{s}^\mathrm{j}_\mathrm{i}) $ consists of bounding box coordinates $ \hat{\bm{b}}^\mathrm{j}_\mathrm{i} $, class label $ \hat{c}^\mathrm{j}_\mathrm{i} $, and confidence score $ \hat{s}^\mathrm{j}_\mathrm{i} $. $\mathrm{N}$ and $\mathrm{N}_{x_{\mathrm{i}}}$ are the number of samples in the dataset and the number of detections for the i-th input image, respectively.

In OD, after initial detections, the Non-Maximum Suppression (NMS) process \cite{ren2015faster} filters out redundant or suboptimal detections. NMS first generates a set of candidates for each detection. These candidates are defined as all other predictions sharing the same class label and having an Intersection-over-Union (IoU), a measure of the overlap between bounding boxes, above a threshold $\epsilon$. Detections below a confidence threshold $\delta$ are then discarded. The candidate set for a given detection $ \hat{\bm{y}}^\mathrm{j}_\mathrm{i} $ is defined as:

\begin{equation}
\mathds{C}_{\hat{\bm{y}}^\mathrm{j}_\mathrm{i}} = \{
        \hat{\bm{y}}^\mathrm{k}_\mathrm{i}  |  \mathrm{k} \neq \mathrm{j} 
        , IoU(\hat{\bm{y}}^\mathrm{k}_\mathrm{i}, \hat{\bm{y}}^\mathrm{j}_\mathrm{i}) \geq \epsilon
        , \hat{c}^\mathrm{k}_\mathrm{i} = \hat{c}^\mathrm{j}_\mathrm{i} 
        , \hat{s}^\mathrm{k}_\mathrm{i} \geq \delta\}.
    \label{eq:candlist}
\end{equation}

After filtering, the remaining detection with the highest confidence score is retained as the final prediction. AP \cite{ren2015faster} has been employed to quantify OD performance.

\subsubsection{Performance Metrics}
D-ECE \cite{kuppers2020multivariate} is just used in our benchmark to study domain shift impacts on the calibration error. D-ECE \cite{kuppers2020multivariate} was calculated by binning the confidence space as well as box coordination parameters space in which there are $\mathrm{N}_\mathrm{k}$ equally distributed bins corresponding to the k-th dimension, ${\displaystyle \mathrm{N}_{total}=\prod_{\mathrm{k}=1}^{\mathrm{K}}\mathrm{N}_\mathrm{k}}$. The goal of binning is to account for variations in calibration error across different confidence levels and spatial dimensions, ensuring that errors are captured in an unbiased manner. Therefore, D-ECE could be formalized
\begin{equation}
    \text{D-ECE}_{\mathrm{k}}=\sum_{\mathrm{n}=1}^{\mathrm{N}_{\mathrm{k}}}\frac{|I(\mathrm{n})|}{|\hat{y}|}|prec(\mathrm{n})-conf(\mathrm{n})|.
    \label{eq:dece}
\end{equation}
Within \cref{eq:dece}, $|I(\mathrm{n})|$ is used to describe the cardinality of the bin, whereas $|\hat{y}|$ represents the total number of detections. $conf(\mathrm{n})$ denotes the mean confidence score of the detections within the bin, whereas $prec(\mathrm{n})$ is a statistical metric that quantifies the proportion of true positives among the detections in the bin.

We utilize MCDO-based and gradient self-information metrics to estimate uncertainty in the presence of DS and compare their capabilities with our proposed method, \textbf{MCDO-map}, to take advantage of them in our UDA method. The utilized methods are referred to as \textbf{MCDO-NMS} and \textbf{Grad-loss}, respectively. The \textbf{Grad-loss} captures the degree of epistemic uncertainty for each detection. \textbf{Grad-loss-localization} and \textbf{Grad-loss-classification} refer to the localization and classification terms, respectively.

The MCDO-based method involves running multiple inference passes with dropout activated. Detections are matched to a candidate list based on the highest IoU threshold. The standard deviation of the localization parameters and the entropy of the mean classification probabilities are calculated for each list. This technique includes \textbf{MCDO-NMS-localization} and \textbf{MCDO-NMS-classification}. For details on these methods, please refer to the supplementary materials.

As opposed to utilizing NMS-based or data association techniques in an MCDO scheme, a score map is constructed in a pixel-wise manner. Given the predictions, we convert the detection outputs to a 3D map. Let $ \hat{\mathds{Y}_\mathrm{i}} = \{\hat{\bm{y}}^{\mathrm{j}}_\mathrm{i}\}_{\mathrm{j}=1}^{\mathrm{N}_{x_{\mathrm{i}}}} $ be the set of detections for an input image. The score map $ \bm{S} $ is a tensor of shape $ (\mathrm{H}, \mathrm{W}, \mathrm{C}+1) $, where $ \mathrm{H} $ and $ \mathrm{W} $ are the height and width of the input image, and $ \mathrm{C} $ is the number of classes. For each detection $ \hat{\bm{y}}^{\mathrm{j}}_\mathrm{i} $, the score $ \hat{s}^{\mathrm{j}}_\mathrm{i} $ is assigned to each pixel inside the bounding box $ \hat{\bm{b}}^{\mathrm{j}}_\mathrm{i} $,
\begin{equation}
    \begin{gathered}
        \bm{S}(x, y, c) \mathrel{+}= 
        \begin{cases} 
             \hat{s}^{\mathrm{j}}_\mathrm{i} & (x, y) \in \hat{\bm{b}}^{\mathrm{j}}_\mathrm{i}, c = \hat{c}^{\mathrm{j}}_\mathrm{i} \\
             (1 - \hat{s}^{\mathrm{j}}_\mathrm{i}) & (x, y) \in \hat{\bm{b}}^{\mathrm{j}}_\mathrm{i}, c = \text{BG}
        \end{cases}.
    \end{gathered}
    \label{eq:scoremap}
\end{equation}
The $\bm{S}$ is zero initiated, resulting in all-zero vectors for pixels that are not contained within any bounding box. For these pixels, we replace the all-zero vectors with a vector with a 1 for the background element and zeros for all other elements.
After populating the score map, we normalize it using the softmax function.
Next, we calculate the mean and standard deviation of the score map over multiple iterations of our object detector forward path that are
\begin{equation}
    \bar{\bm{S}} = \frac{1}{\mathrm{N_{map}}} \sum_{\mathrm{n}=1}^{\mathrm{N_{map}}} \bm{S}_\mathrm{n}, 
    \sigma_{\bm{S}} = \sqrt{\frac{1}{\mathrm{N_{map}}} \sum_{\mathrm{n}=1}^{\mathrm{N_{map}}} (\bm{S}_\mathrm{n} - \bar{\bm{S}})^2}.
    \label{eq:mapstdent}
\end{equation}
Finally, we compute the entropy of the mean score map, $ H_{\bar{\bm{S}}}(x, y, :) = -\sum_{c} \bm{S}(x, y, c) \log \bm{S}(x, y, c) $, and concatenate the standard deviations to create the uncertainty map, $ \textbf{MCDO-map} = concat(H_{\bar{\bm{S}}}(x, y, :), \sum_{c} \sigma_{\bm{S}}(x, y, c)) $.
From an intuitive standpoint, it can be observed that increasing changes in localization parameters of the predictions are associated with a corresponding increase in the standard deviation of the boundaries surrounding pixels. For example, in \cref{fig:uda}, the left magnified detection in the target std map shows higher deviation (with colors closer to red) compared to the source std map, where the corresponding detection is mostly blue, indicating lower deviation across pixels.
Similarly, a higher frequency of change in prediction scores is shown to be linked to an elevated level of entropy. The same behavior in the entropy maps can be observed in \cref{fig:uda}.
In contrast to traditional MCDO-based approaches, instead of handling individual bounding boxes from each iteration and suffering postprocessing complexity \cite{feng2021review}, our method generates a pixel-wise score map during each iteration and avoids complex postprocessing.
\subsubsection{Domain Shift and Adaptation}
DS occurs when the training (source) domain $ \mathcal{D}^s $ and the testing (target) domain $ \mathcal{D}^t $ differ, leading to a performance drop in machine learning models. Let $ \mathcal{D}^s = \{(\bm{X}^{s}_{\mathrm{i}}, \mathds{Y}^{s}_{\mathrm{i}})\}_{\mathrm{i}=1}^{\mathrm{N}^s} $ and $ \mathcal{D}^t = \{(\bm{X}^{t}_{\mathrm{i}}, \mathds{Y}^{t}_{\mathrm{i}})\}_{\mathrm{i}=1}^{\mathrm{N}^t} $. We denote the source and target distributions as $ p^s(\bm{X}, \mathds{Y}) $ and $ p^t(\bm{X}, \mathds{Y}) $, respectively. Distribution shift is defined as $ p^s(\bm{X}, \mathds{Y}) \neq p^t(\bm{X}, \mathds{Y}) $. If we consider $ p^s(\bm{X}, \mathds{Y}) = p^s(\bm{X})p^s(\mathds{Y} | \bm{X}) $ and $ p^t(\bm{X}, \mathds{Y}) = p^t(\bm{X})p^t(\mathds{Y} | \bm{X}) $ , the DS happens when $p^s(\bm{X}) \neq p^t(\bm{X}), p^s(\mathds{Y} | \bm{X}) = p^t(\mathds{Y} | \bm{X}).$

Our UDA method focuses on leveraging uncertainty information to enhance the robustness of the object detector in the presence of DS. We got inspired by ADVENT \cite{vu2019advent} while modifying it by changing the representation of input data to the discriminator and introducing a novel uncertainty estimation method.
The intuition behind this method is that DS introduces uncertainty in predictions, especially in regions where the model is less confident. Our uncertainty maps highlight areas where the domain shift has the most impact, guiding the adaptation process to focus on these challenging regions.
Following the concatenation process, the \textbf{MCDO-map} is subsequently forwarded to a domain discriminator to fool it, initiating adversarial training (\cref{fig:uda}). The calculation of the overall loss is $\mathcal{L}_{total} = \mathcal{L}_{detection} - \lambda \times \mathcal{L}_{adv},$
where the detection loss $ \mathcal{L}_{detection} $ is a combination of cross-entropy classification and smooth $ L_1 $ regression loss, $ \mathcal{L}_{detection} = \mathcal{L}_{cls} + \mathcal{L}_{reg} $. The adversarial loss $ \mathcal{L}_{adv} $ is
\begin{equation}
    \begin{gathered}
        \mathcal{L}_{adv} = -\mathbb{E}_{\bm{X} \sim p^s} \log D(\text{MCDO-map}(\bm{X})) 
        \\
        - \mathbb{E}_{\bm{X} \sim p^t} \log (1 - D(\text{MCDO-map}(\bm{X}))),
    \end{gathered}
    \label{eq:advloss}
\end{equation}
when the $D$ is the discriminator network. The detection base network is updated to minimize the total loss, $ \bm{\omega}^* = \arg \min_{\bm{\omega}} \mathcal{L}_{total} $, while the discriminator network is updated to maximize the adversarial loss, $ \theta^* = \arg \max_{\theta} \mathcal{L}_{adv} $. 
\subsection{Benchmark Scenarios}
We will begin our benchmark with \cref{tab:shiftresults}, illustrating the impact of domain shift on object detection using the AP, uncertainty, and D-ECE metrics. 
\cref{tab:unccomp} presents a comparison between our proposed MCDO-map method and other uncertainty metrics. 
Finally, our novel uncertainty-aware UDA object detector is compared with SOTA UDA methods in \cref{tab:udares}. Supplementary materials have been provided to support our discussions.

\textbf{Background-wise Metrics:} To assess OD performance under BG shifts, we introduce BG-wise metrics, which calculate metrics separately for different BGs (e.g., sky, tree, ground). Given detections $\hat{\mathds{Y}}$ and ground truth $\mathds{Y}$, we classify each into BG categories using segmentation maps to identify the background category to which most of the pixels within the bounding box belong. The metric $M$ for each BG category can be expressed as $M_{\text{bg}}(\hat{\mathds{Y}}, \mathds{Y}) = M(\hat{\mathds{Y}}_{\text{bg}}, \mathds{Y}_{\text{bg}}), \text{bg}\in\{\text{sky, tree, ground}\}.$
These metrics provide a detailed analysis of how different BGs affect object detection performance.
\subsection{Experiments and Results}
\label{subsec:exp_res}

\begin{table*}
    \centering
    {\small{
    \begin{tabular}{@{}l|cccc|llll|l|llll@{}}
        \toprule
        \parbox[t]{2mm}{\multirow{5}{*}{\rotatebox[origin=c]{90}{\textbf{Source Domain}}}} & \multicolumn{4}{c|}{Target Domain} & \multicolumn{4}{c|}{$AP_{50}\uparrow\%$} & MCDO-map$\downarrow$ & \multicolumn{4}{c}{D-ECE$\downarrow\times10^{-6}$} \\
        \cmidrule{2-9}
        \cmidrule{11-14}
        & PoV & Source & Season & Weather & Total & Sky & Tree & Ground & $\times10^{-4}$ & Total & Sky & Tree & Ground \\
        \cmidrule{2-14}
        \parbox[t]{2mm}{\multirow{15}{*}{\rotatebox[origin=c]{90}{\textbf{I}}}} & ground & synthetic & winter & normal & 40.7 & 67.1 & 0.2 & 54.7 & 3582 & 674 & 248 & 576 & 1176\\
         \cmidrule{2-14}
        & ground & \cellcolor{lightgray}real & winter & normal
        & 3.2 & 9.5 & 0.1 & 0.0 & 5355 & 2253 & 1586 & 69 & 2623 \\
        \midrule
        & ground & synthetic & adverse & \cellcolor{lightgray} rainy
        & - & 75.8 & - & - & 3866 & - & 7679 & - & - \\
        & ground & synthetic & adverse & \cellcolor{lightgray} snowy
        & - & 63.0 & - & - & 4686 & - & 6803 & - & - \\
        & ground & synthetic & adverse & \cellcolor{lightgray} foggy
        & - & 98.7 & - & - & 4454 & - & 1796 & - & - \\
        \cmidrule{2-14}
        & \cellcolor{lightgray}aerial & synthetic & winter & normal
        & 12.7 & 35.6 & 0.0 & 2.4 & 4287 & 4095 & 2063 & 2819 & 5518 \\
        & \cellcolor{lightgray}aerial & synthetic & \cellcolor{lightgray}fall & normal
        & 41.5 & 81.0 & 10.8 & 32.9 & 4680 & 7124 & 8092 & 5275 & 3022 \\
        & \cellcolor{lightgray}aerial & synthetic & \cellcolor{lightgray}summer & normal
        & 51.5 & 74.3 & 64.8 & 15.3 & 4677 & 6447 & 7327 & 2245 & 1444 \\
        \cmidrule{2-14}
        & \cellcolor{lightgray}aerial & synthetic & adverse & \cellcolor{lightgray}rainy
        & - & 26.1 & - & - & 4084 & - & 2923 & - & - \\
        & \cellcolor{lightgray}aerial & synthetic & adverse & \cellcolor{lightgray}snowy
        & - & 19.8 & - & - & 4835 & - & 8008 & - & - \\
        & \cellcolor{lightgray}aerial & synthetic & adverse & \cellcolor{lightgray}foggy
        & - & 35.7 & - & - & 4394 & - & 4142 & - & - \\
        \midrule
        \parbox[t]{2mm}{\multirow{1}{*}{\rotatebox[origin=c]{90}{\textbf{II}}}} & ground & real & winter & normal & 34.4 & 73.7 & 0.5 & 29.0 & 2673 & 1242 & 1105 & 1694 & 0329\\
        \bottomrule
    \end{tabular}
    }}
    \caption{DS impact on Faster R-CNN detector trained on source domain I or II in terms of AP, uncertainty, and D-ECE. The Faster R-CNN detector is validated on different target domains. Shaded cells show the shifted element in each target domain. AP and D-ECE are reported BG-wise, although the MCDO-map uncertainty is reported totally. For adverse weather, we have only the sky background. Source domain I: ground-synthetic-winter-normal-sky, source domain II: ground-real-winter-normal-sky}
    \label{tab:shiftresults}
\end{table*}
\begin{figure}
    \centering
    \includegraphics[width=\linewidth, height=0.7\linewidth]{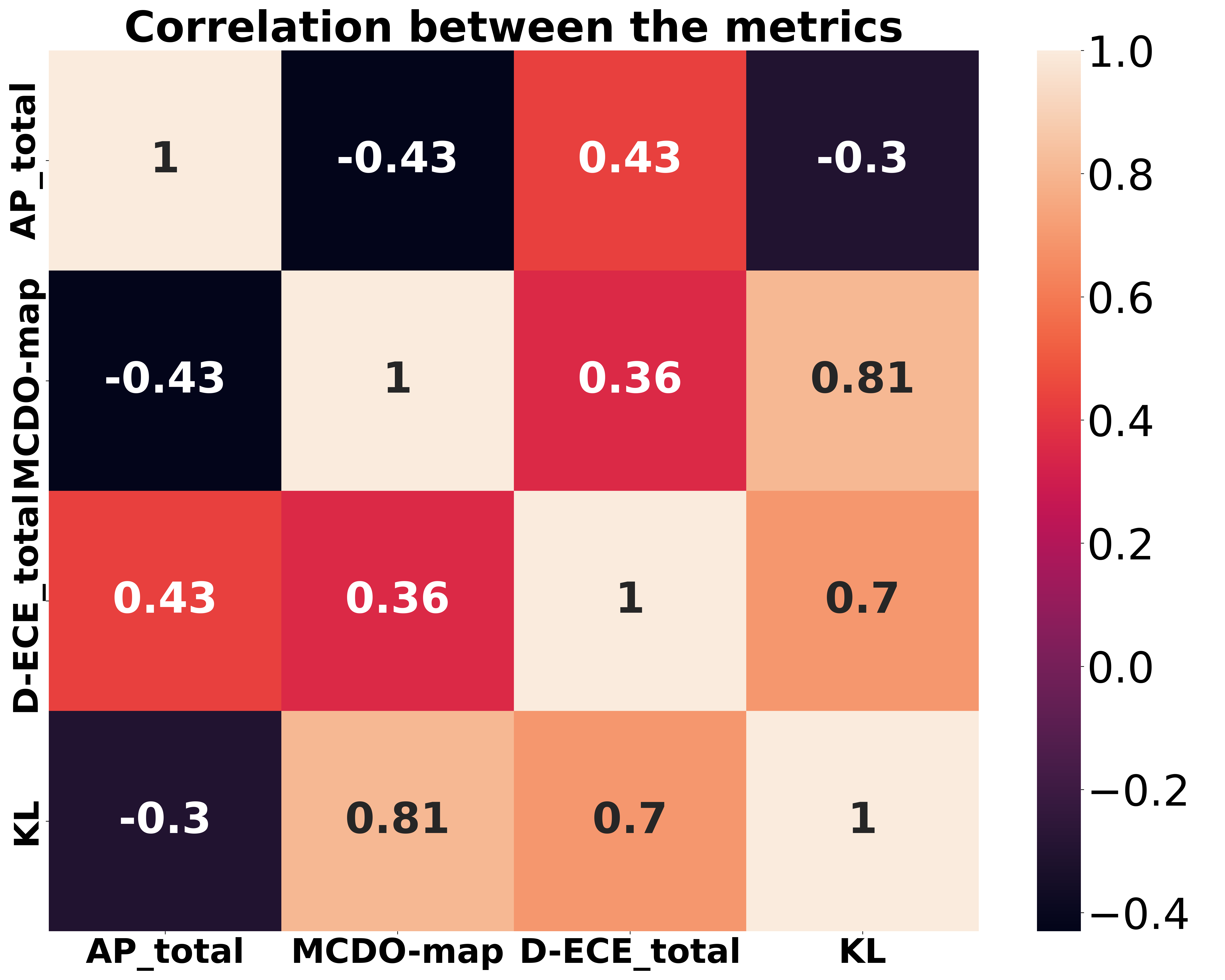}
    \caption{Correlation heatmap: The KL divergence of feature maps distributions and all metrics are calculated between the source (domain I in \cref{tab:shiftresults}) and target domains. MCDO-map's high positive correlation with KL divergence shows high capabilities of MCDO-map to capture the DS.}
    \label{fig:corr}
\end{figure}
\cref{tab:shiftresults} shows different DS scenarios and their impact on object detection models. It is our contribution that metrics are reported background-wise, highlighting the influence of background shifts. The AP under the Sky column in row one is the reference. Significant AP decreases are evident, such as in row two when shifting from synthetic to real. PoV and weather shifts in rows 9 to 11 also show notable changes. Comparing the reference AP and the sky APs in other rows demonstrates decreased APs for tree and ground backgrounds (see Fig. 7 in supplementary material). The domain I in \cref{tab:shiftresults} is the source domain all over the text unless other domains are mentioned.


We calculated the Kullback-Leibler (KL) divergence between source and target domain feature map distributions (see supplementary material)
to analyze the relationship between metrics and different shifts from \cref{tab:shiftresults}. \cref{fig:corr} shows a heatmap of Pearson correlations among AP, D-ECE, MCDO-map, and KL divergence. The high positive correlation between our MCDO-map and KL divergence indicates the MCDO-map's effectiveness in capturing DS.

The negative correlation of AP with MCDO-map and KL divergence suggests that higher AP corresponds to lower uncertainty and smaller feature map distribution distances. The positive correlation between AP and D-ECE indicates model miscalibration under DS. Additionally, the positive relation between D-ECE, MCDO-map, and KL divergence highlights their significant association with DS.
\begin{table*}
    \centering
    {\small{
    \begin{tabular}{cccc|lllllll|l|ll}
        \toprule
        \multicolumn{4}{c|}{Validation Domain} & \multicolumn{7}{c|}{MCDO-NMS$\times10^{-3}$} & MCDO-Map & \multicolumn{2}{c}{grad-loss$\times10^{-3}$} \\
        \multicolumn{4}{c|}{} & \multicolumn{3}{c}{Localization} & & \multicolumn{3}{c|}{Classification} & $\times10^{-4}$ & Loc. & Cls. \\
        \cmidrule{1-7}
        \cmidrule{9-14}
        PoV & Source & Season & Weather & Total & TP & FP && Total & TP & FP &  &  & \\
        \midrule
        ground & synthetic & winter & normal & 107 & 63 & 107 &&  436 & 141 & 439 & 3582 & 445 & 564 \\
        \midrule
        ground & \cellcolor{lightgray}real & winter & normal & 173 & 56 & 180 && 495 & 298 & 508 & 5355 & 434 & 692 \\
        \midrule
        ground & synthetic & adverse & \cellcolor{lightgray}rainy &  98 & 67 & 99 &&  401 & 112 & 404 & 3866 & 390 & 634 \\
        ground & synthetic & adverse & \cellcolor{lightgray}snowy &  79 & 58 & 79 &&  454 & 183 & 456 & 4686 & 458 & 658 \\
        ground & synthetic & adverse & \cellcolor{lightgray}foggy &  85 & 83 & 85 &&  449 & 241 & 484 & 4454 & 335 & 642 \\
        \midrule
        \cellcolor{lightgray}aerial & synthetic & winter & normal &  101 & 62 & 103 &&  407 & 166 & 417 & 4287 & 444 & 500 \\
        \cellcolor{lightgray}aerial & synthetic & \cellcolor{lightgray}fall & normal &  92 & 105 & 92 &&  523 & 312 & 529 & 4680 & 435 & 653 \\
        \cellcolor{lightgray}aerial & synthetic & \cellcolor{lightgray}summer & normal & 94 & 92 & 94 && 538 & 269 & 543 & 4677 & 434 & 658 \\
        \midrule
        \cellcolor{lightgray}aerial & synthetic & adverse & \cellcolor{lightgray}rainy &  100 & 092 & 101 &&  440 & 238 & 446 & 4084 & 468 & 554 \\
        \cellcolor{lightgray}aerial & synthetic & adverse & \cellcolor{lightgray}snowy &  85 & 67 & 85 &&  598 & 298 & 605 & 4835 & 445 & 690 \\
        \cellcolor{lightgray}aerial & synthetic & adverse & \cellcolor{lightgray}foggy &  105 & 116 & 103 &&  468 & 390 & 477 & 4394 & 445 & 635 \\
        \bottomrule
    \end{tabular}
    }}
    \caption{Uncertainty Metrics Comparison: The network is Faster R-CNN trained on the \cref{tab:shiftresults}'s domain I. Each row shows the validation domain in which experiments have been done for the sky background. In each row, the uncertainty level was evaluated using three different methods. MCDO-NMS reported separately for TP and FP detections. Shaded cells show the shifted element. Loc.: localization, Cls.: Classification}
    \label{tab:unccomp}
\end{table*}

The goal of \cref{tab:unccomp} is to compare our MCDO-map method with the MCDO-NMS and Grad-loss. Our method consistently shows increased uncertainty with DS, highlighting its effectiveness in capturing DSs. The wider violin plots for the MCDO-map in Fig. 4 of supplementary material demonstrate its superior capability to separate different DS levels compared to other metrics. MCDO-NMS-Classification for TPs shows some capability in separating different DSs, highlighted in \cref{tab:unccomp} and supplementary material, but requires supervision and often decreases with DS. Grad-loss-localization is consistently capturing the DS (\cref{tab:unccomp})  but lacks the potential to separate DSs effectively.

Our results in \cref{fig:corr} and supplementary material further support the MCDO-map's effectiveness in capturing DS. Thus, we conclude that the MCDO-map is the best method for our UDA approach, offering significant improvements over traditional uncertainty estimation techniques.
\begin{table*}
    \centering
    \begin{subtable}{0.33\linewidth}
        \centering
        {\small{
        \begin{tabularx}{\linewidth}{lXXXX}
            \toprule
            Method & Total & Sky & Tree & Gr. \\
            \midrule
            S.O. & 40.7 & \textbf{67.1} & 0.2 & 54.7 \\
            \midrule
            C.fMix \cite{mattolin2023confmix}  & 44.0  & 66.8 & 5.4 & \textbf{56.3} \\
            SAPN \cite{zheng2022sapnet} & 41.6  & 61.4 & 5.1 & 53.2 \\
            PT \cite{chen2022learning} & 42.5  & 64.9 & 6.3 & 55.3 \\
            \midrule
            \textbf{\textit{Ours}} & \textbf{46.3} & 62.1 & \textbf{10.7} & 55.9 \\
            \bottomrule
        \end{tabularx}
        }}
        \caption{ground-synthetic-winter-normal-tree}
    \end{subtable}
    \hfill
    \begin{subtable}{0.33\linewidth}
        \centering
        {\small{
        \begin{tabularx}{\linewidth}{lXXXX}
            \toprule
            Method & Total & Sky & Tree & Gr. \\
            \midrule
            S.O. & 40.7 & \textbf{67.1} & 0.2 & 54.7 \\
            \midrule
            C.fMix \cite{mattolin2023confmix}  & 42.5 & 64.5 & 0.5 & 56.0 \\
            SAPN \cite{zheng2022sapnet} & 41.6  & 61.4 & 5.1 & 53.2 \\
            PT \cite{chen2022learning} & 42.0 & 61.5 & 0.9 & 54.2 \\
            \midrule
            \textbf{\textit{Ours}} & \textbf{44.8} & 66.0 & \textbf{1.2} & \textbf{56.5} \\
            \bottomrule
        \end{tabularx}
        }}
        \caption{ground-synthetic-winter-normal-ground}
    \end{subtable}
    \hfill
    \begin{subtable}{0.33\linewidth}
        \centering
        {\small{
        \begin{tabularx}{\linewidth}{lXXXX}
            \toprule
            Method & Total & Sky & Tree & Gr. \\
            \midrule
            S.O. & 12.7 & 35.6 & 0.0 & 2.4 \\
            \midrule
            C.Mix \cite{mattolin2023confmix} & 15.3 & 37.1 & 0.4 & 3.2 \\
            SAPN \cite{zheng2022sapnet} & 14.3 & 35.9 & 0.3 & 2.9 \\
            PT \cite{chen2022learning} & 14.7 & 36.6 & \textbf{0.6} & 3.1 \\
            \midrule
            \textbf{\textit{Ours}} & \textbf{17.8} & \textbf{41.3} & 0.5 & \textbf{3.6} \\
            \bottomrule
        \end{tabularx}
        }}
        \caption{aerial-synthetic-winter-normal}
    \end{subtable}
    \hfill
    \begin{subtable}{0.33\linewidth}
        \centering
        {\small{
        \begin{tabularx}{\linewidth}{lXXXX}
            \toprule
            Method & Total & Sky & Tree & Gr. \\
            \midrule
            S.O & 3.2 & 9.5 & 0.1 & 0.0 \\
            \midrule
            C.Mix \cite{mattolin2023confmix} & 5.0 & 13.2 & 0.7 & 0.3 \\
            SAPN \cite{zheng2022sapnet} & 4.8 & 12.8 & 0.6 & 0.2 \\
            PT \cite{chen2022learning} & 4.6 &12.5 & 0.5 & 0.2 \\
            \midrule
            \textbf{\textit{Ours}} & \textbf{5.7} & \textbf{14.0} & \textbf{0.8} & \textbf{0.4} \\
            \bottomrule
        \end{tabularx}
        }}
        \caption{ground-real-winter-normal-sky}
    \end{subtable}
    \hfill
    \begin{subtable}{0.33\linewidth}
        \centering
        {\small{
        \begin{tabularx}{\linewidth}{lXXXX}
            \toprule
            Method & Total & Sky & Tree & Gr. \\
            \midrule
            S.O. & 3.2 & 9.5 & 0.1 & 0.0 \\
            \midrule
            C.Mix \cite{mattolin2023confmix} & 7.0 & 10.5 & 5.0 & 0.3 \\
            SAPN \cite{zheng2022sapnet} & 6.8 & 10.0 & 4.8 & 0.2 \\
            PT \cite{chen2022learning} & 6.5 & 10.2 & 4.6 & 0.3 \\
            \midrule
            \textbf{\textit{Ours}} & \textbf{8.2} & \textbf{11.0} & \textbf{6.0} & \textbf{0.5} \\
            \bottomrule
        \end{tabularx}
        }}
        \caption{ground-real-winter-normal-tree}
    \end{subtable}
    \hfill
    \begin{subtable}{0.33\linewidth}
        \centering
        {\small{
        \begin{tabularx}{\linewidth}{lXXXX}
            \toprule
            Method & Total & Sky & Tree & Gr. \\
            \midrule
            S.O. & 3.2 & 9.5 & 0.1 & 0.0 \\
            \midrule
            C.Mix \cite{mattolin2023confmix} & 7.1 & \textbf{10.0} & 4.8 & 0.4 \\
            SAPN \cite{zheng2022sapnet} & 6.8 & 9.8 & 4.9 & 0.2 \\
            PT \cite{chen2022learning} & 7.1 & 9.7 & \textbf{5.2} & 0.3 \\
            \midrule
            \textbf{\textit{Ours}} & \textbf{8.9} & 9.9 & 4.6 & \textbf{0.9} \\
            \bottomrule
        \end{tabularx}
        }}
        \caption{ground-real-winter-normal-ground}
    \end{subtable}
    \caption{Background-wise mAP of different UDA methods for different domain shifts: Our UDA method surpasses the SOTA techniques in most cases. The target domain is written below each subtable. S.O.: Source Only (trained network on \cref{tab:shiftresults}'s domain I).}
    \label{tab:udares}
\end{table*}

In \cref{tab:udares}, the results of some SOTA UDA object detectors on the DrIFT dataset are reported alongside our results. One significant DS is from sky to tree, where AP dropped from 67.1 to 0.2 (\cref{tab:shiftresults}, first row). Our UDA method outperforms others with an AP of 10.7 for the tree background and 46.3 in total, demonstrating its effectiveness in adapting to different BGs. Similarly, for the ground background domain, our method achieves an AP of 44.8 in total and 1.2 for the tree background, showcasing robustness.

For the aerial-synthetic-winter-normal domain, multiple BGs in each frame could take place, so we did not specify any BG for it. Our UDA method achieves the highest AP in total (17.8), sky (41.3), and tree (0.5), indicating its capability to adapt to different PoVs and complex scenes. In the ground-real-winter-normal-sky domain, our method achieves the highest AP in total (5.7) and tree (0.8), proving its effectiveness with real-world data and different seasons. The results are consistent for tree and ground backgrounds as well, demonstrating our method's adaptability for two types of DSs occurring simultaneously.

Our method aims to deceive a domain discriminator by making the uncertainty maps for both source and target domains nearly identical. This dual focus on source and target domain alignment is crucial for robust performance across various DSs. The adaptation process involves a trade-off, accepting some degradation in the source domain to achieve significant improvements in the target domain.

\section{Conclusion}

The DrIFT dataset addresses the need for a vast study of domain shift in drone detection by introducing fourteen distinct domains and emphasizing background shift utilizing background segmentation maps. Our findings show a positive correlation between MCDO-map uncertainty, domain shift, and D-ECE, and a negative correlation with AP. The MCDO-map outperformed other uncertainty metrics in capturing domain shift in the DrIFT dataset. Our uncertainty-aware UDA on object detection also surpassed SOTA methods in the DrIFT dataset. In the future, we aim to explore more nuanced domain adaptation techniques that minimize the source domain performance degradation, which is a drawback for our UDA method in some cases.


{\small
\bibliographystyle{ieee_fullname}
\bibliography{WACV2025_DrIFT_arxiv}
}
\newpage
\clearpage
\renewcommand{\thesection}{S\arabic{section}}
\setcounter{section}{0}
\setcounter{figure}{0}
\setcounter{table}{0}
\setcounter{equation}{0}
\renewcommand{\thefigure}{S\arabic{figure}}
\renewcommand{\thetable}{S\arabic{table}}
\renewcommand{\theequation}{S\arabic{equation}}

\section{Introduction}
\label{supp_sec:intro}
This supplementary material contains important information that could not be included in the main paper due to space constraints and aims to support the discussions in the main paper. The structure of the main paper is followed.

\section{DrIFT Dataset}
\label{supp_sec:drift}
\subsection{Dataset Characteristics and Statistics}
\label{supp_subsec:stats}
\begin{figure*}
  \centering
  \includegraphics[width=\linewidth]{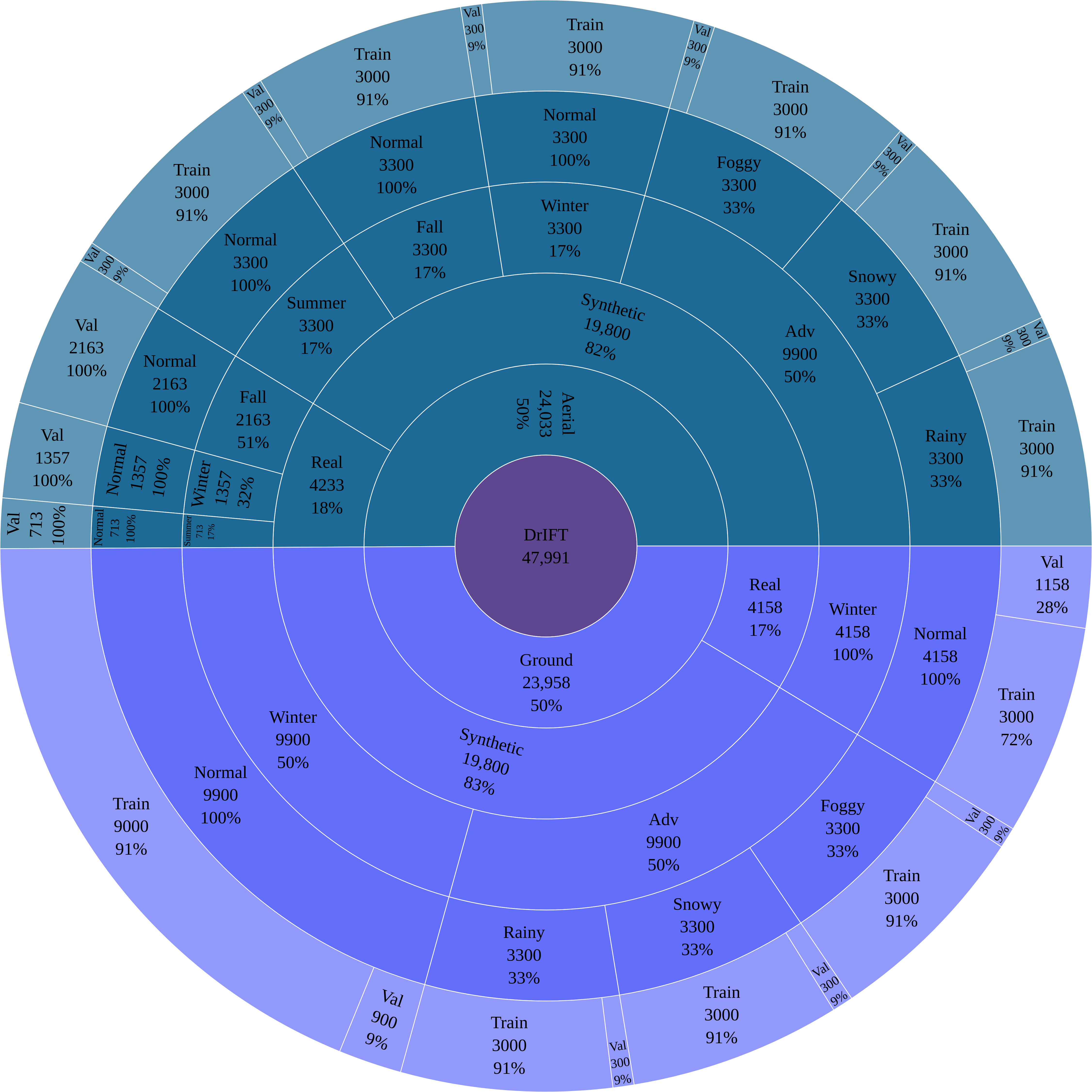}
  \caption{Hierarchical sunburst chart of the DrIFT dataset: The DrIFT dataset contains aerial and ground views in real-world and simulated environments. There are numerous domains based on the various seasons and weather. The chart displays the number and percentage of the samples within the parent category. Adv: adverse}
  \label{supp_fig:datasetsunburst}
\end{figure*}
\begin{figure*}
    \centering
    \includegraphics[width=\linewidth]{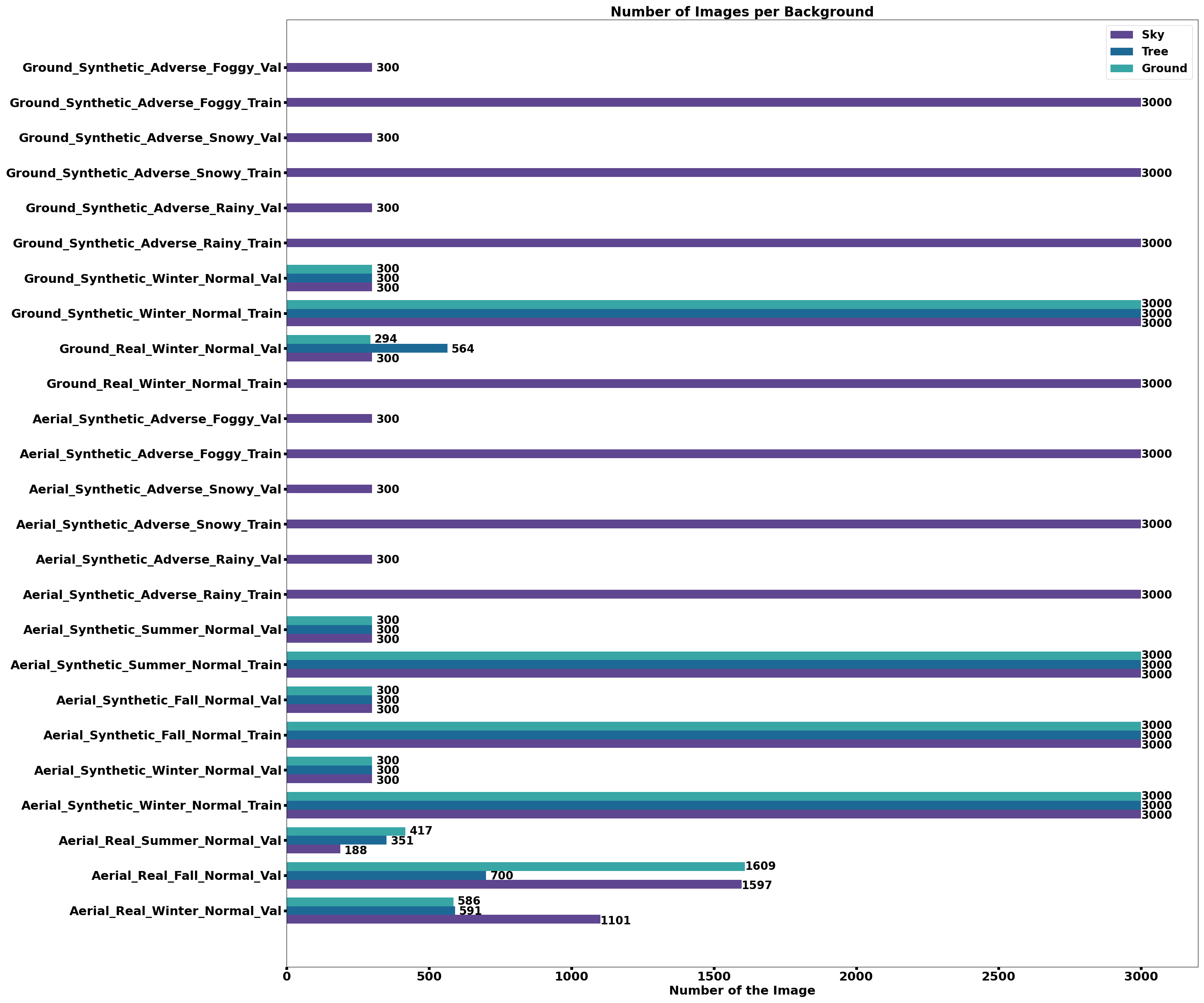}
    \caption{Number of existing background samples in each domain: We aim to maintain an equal number of background samples for each domain's training and validation sets, unless the distribution of real data prevents us from adhering to this guideline.}
    \label{supp_fig:background}
\end{figure*}

Fig. 1 in the main paper displays a variety of backgrounds from our dataset, including the sky, trees, and ground during three distinct seasons (fall, winter, and summer) or adverse weather conditions (foggy, snowy, and rainy). \cref{supp_fig:datasetsunburst} demonstrates that the DrIFT possesses $47,991$ image frames.
As discussed in Subsec. 3.1 "The DrIFT Story" in the main paper, we attempted to keep the balance between training and validation sets for almost all domains: $3,000$ frames for training and 300 frames for validation.
This standard practice facilitates a proper platform for evaluating the UDA algorithms. \cref{supp_fig:background} shows the number of existing background samples in each domain. It is important to note that, as shown in the last three rows of \cref{supp_fig:background}, the dataset includes only a validation set for the aerial-real domains, without a corresponding training set. 
Additionally, it is noteworthy that our adverse weather domains only contain a sky background. Hence we have avoided reporting metrics for tree and ground backgrounds within these domains in \cref{supp_tab:unccomp}.

\begin{figure*}
  \centering
  \begin{subfigure}{0.24\linewidth}
    \includegraphics[width=\linewidth,height=\linewidth]{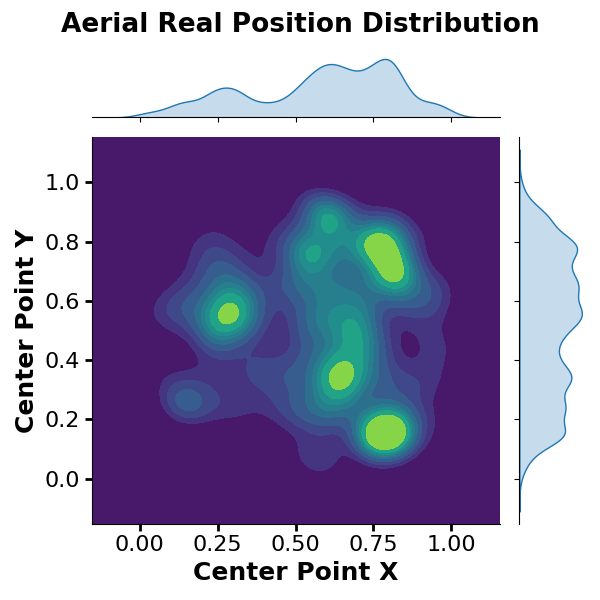}
    \caption{}
    \label{supp_fig:Loc_Size-a}
  \end{subfigure}
  \begin{subfigure}{0.24\linewidth}
    \includegraphics[width=\linewidth,height=\linewidth]{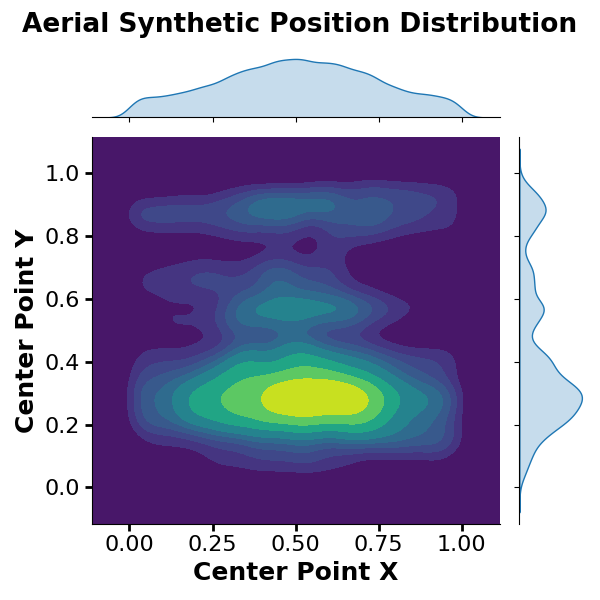}
    \caption{}
    \label{supp_fig:Loc_Size-b}
  \end{subfigure}
  \begin{subfigure}{0.24\linewidth}
    \includegraphics[width=\linewidth,height=\linewidth]{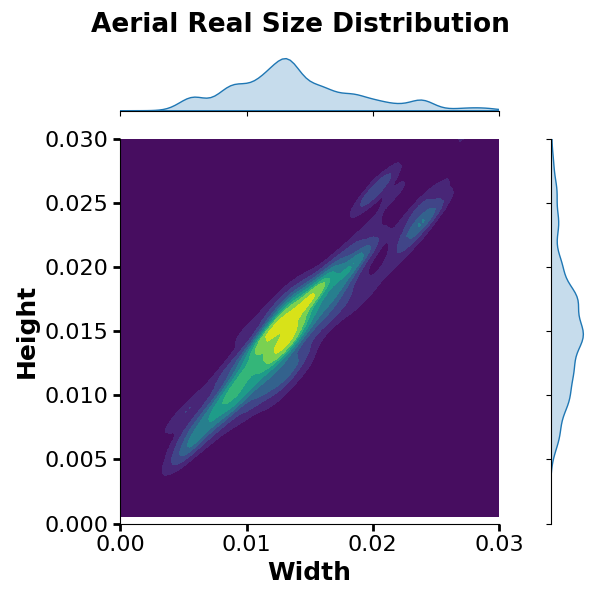}
    \caption{}
    \label{supp_fig:Loc_Size-c}
  \end{subfigure}
  \begin{subfigure}{0.24\linewidth}
    \includegraphics[width=\linewidth,height=\linewidth]{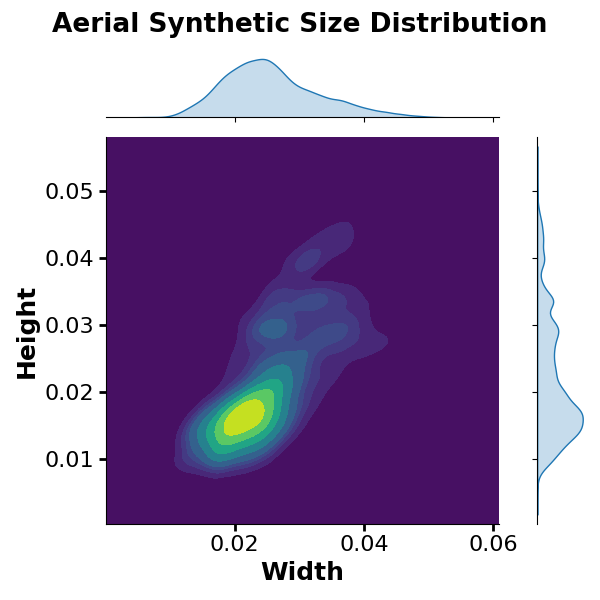}
    \caption{}
    \label{supp_fig:Loc_Size-d}
  \end{subfigure}
  \begin{subfigure}{0.24\linewidth}
    \includegraphics[width=\linewidth,height=\linewidth]{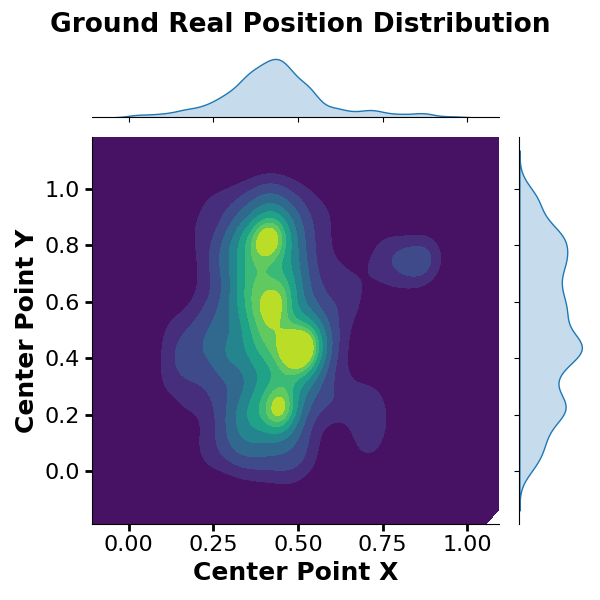}
    \caption{}
    \label{supp_fig:Loc_Size-e}
  \end{subfigure}
  \begin{subfigure}{0.24\linewidth}
    \includegraphics[width=\linewidth,height=\linewidth]{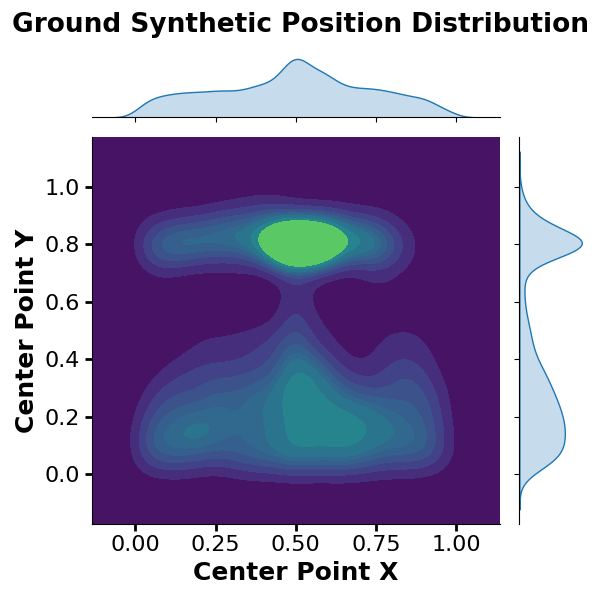}
    \caption{}
    \label{supp_fig:Loc_Size-f}
  \end{subfigure}
  \begin{subfigure}{0.24\linewidth}
    \includegraphics[width=\linewidth,height=\linewidth]{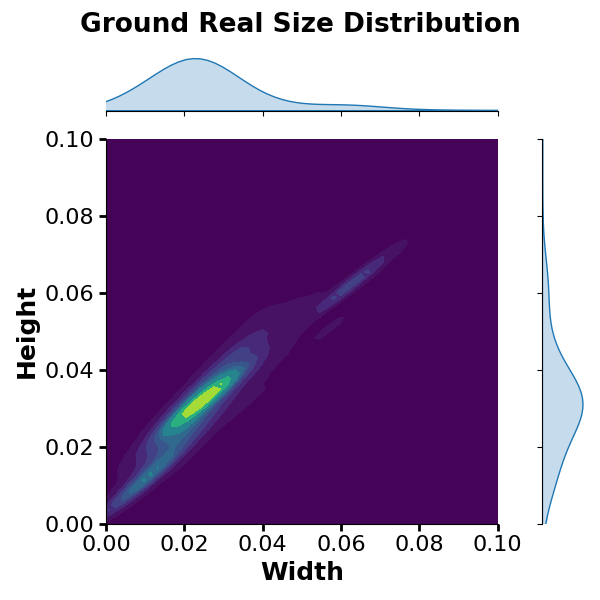}
    \caption{}
    \label{supp_fig:Loc_Size-g}
  \end{subfigure}
  \begin{subfigure}{0.24\linewidth}
    \includegraphics[width=\linewidth,height=\linewidth]{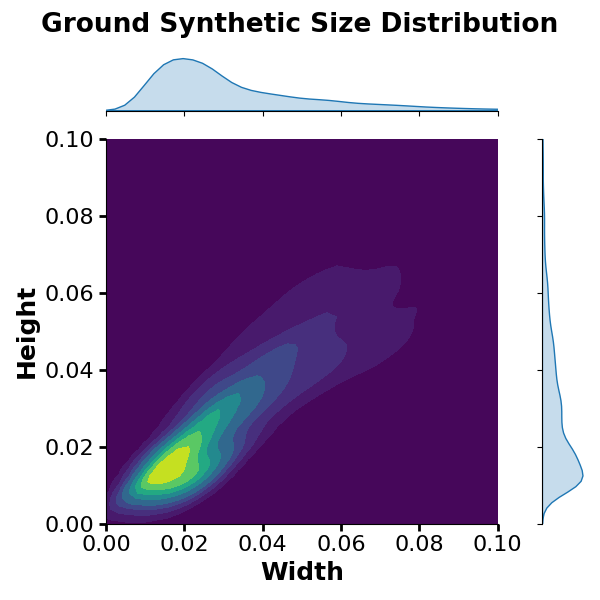}
    \caption{}
    \label{supp_fig:Loc_Size-h}
  \end{subfigure}
  \hfill
  \caption{Center point $(x,y)$, height, and width distributions through training sets for both aerial and ground view with a separation for both real and synthetic parts of our dataset. The height and width distributions show the relatively small size of the objects available in DrIFT.}
  \label{supp_fig:Loc_Size}
\end{figure*}

\cref{supp_fig:Loc_Size} depicts drones' relative size and location distribution in real and synthetic domains. The center point, width, and height are normalized to the image width and height. For the aerial-real data in \cref{supp_fig:Loc_Size-c}, the means of the relative width and height are approximately 0.015, whereas in \cref{supp_fig:Loc_Size-d}, representing aerial-synthetic data, these values are around 0.02 and 0.015, respectively. The width and height of the ground-real, illustrated in \cref{supp_fig:Loc_Size-g}, are about 0.03, although for the ground-synthetic shown in \cref{supp_fig:Loc_Size-h}, these are approximately 0.02. These numbers indicate that we deal with extremely small objects in comparison to other applications, \eg, autonomous land vehicles \cite{Cordts2016Cityscapes, sun2022shift}. It makes DrIFT more challenging in terms of training the detector models.

\section{DrIFT Benchmark}
\label{supp_sec:bench}
\subsection{Methodology}
\label{supp_subsec:meth}

\subsubsection{Uncertainty Estimation}
In \cite{riedlinger2023gradient, riedlinger2022uncertainty}, the researchers employed a gradient function and introduced the concept of self-learning gradient as a metric to evaluate the uncertainty of each detection. If we consider the supervised learning scenario the gradient of the loss function for each detection is $ g(\bm{X}_\mathrm{i}, \bm{y}^\mathrm{j}_\mathrm{i}) = \nabla_{\bm{\omega}} \mathcal{L}(\hat{\bm{y}}^\mathrm{j}_\mathrm{i}, \bm{y}^\mathrm{j}_\mathrm{i})  $. The $ \bm{\omega} $ is the network's weight vector. If the ground truth, $ \bm{y}^\mathrm{j}_\mathrm{i} $, is replaced with detection, $\hat{\bm{y}}^\mathrm{j}_\mathrm{i} $, and the detection is replaced with its candidates, $\hat{\bm{y}}^\mathrm{k}_\mathrm{i}$, the self-gradient metric would be
\begin{equation}
  g^{cand}(\bm{X}_\mathrm{i}, \hat{\bm{y}}^\mathrm{j}_\mathrm{i})= \sum_{\hat{\bm{y}}^\mathrm{k}_\mathrm{i} \in \mathds{C}_{\hat{\bm{y}}^\mathrm{j}_\mathrm{i}}} \nabla_{\bm{\omega}} \mathcal{L}(\hat{\bm{y}}^\mathrm{k}_\mathrm{i}, \hat{\bm{y}}^\mathrm{j}_\mathrm{i}).
  \label{supp_eq:candgrad}
\end{equation}
The self-gradient metric, $ g^{cand}(.) $, referred to as \textbf{Grad-loss}, operates as a characteristic that signifies the degree of epistemic uncertainty, which is the focal point for investigating DS. \textbf{Grad-loss-localization} is called the corresponding localization term of the loss, although \textbf{Grad-loss-classification} points to the classification term in the loss. Nevertheless, it does not inherently encompass the true essence of uncertainty. To consider other methods, we employ a technique based on MC-dropout to capture the inherent uncertainty associated with each detection. In this approach, we activate dropout at inference time and run our model for $\mathrm{N}_{mcdo}$ times. Let us consider the output of the model at each iteration $ \hat{\mathds{Y}}^{\mathrm{m}}_{\mathrm{i}} $  where $\mathrm{m}\in\{1,\dots, \mathrm{N}_{mcdo}\}$. Initially, we create an $ |\hat{\mathds{Y}}^{1}_{\mathrm{i}}| $-length list corresponding to all output detections in the first iteration. Subsequently, we perform some NMS like the one in Eq. 1 of the main paper to have a candidate list and assign the best candidate with the highest $IoU$ to each list. A detection in each iteration is only allowed to be a member of one list, and a new list is created if there is no option with a higher $IoU$ threshold.
Ultimately, we calculate the standard deviation of localization parameters $\sigma_{\bm{b}}^\mathrm{q}$ and the entropy of the mean of the classification probability vector $H_{cls}^\mathrm{q}$, $ \mathrm{q}\in\{1,\dots,N_{mcdo\_out}\} $, respectively. If we assume we have $ N_{mcdo\_out} $ lists of outputs, we can compute the uncertainty for each list as follows:
\begin{equation}
    \sigma_b^\mathrm{q}=\sqrt{\frac{1}{\mathrm{c}_\mathrm{q}}\sum_{\mathrm{n}=1}^{\mathrm{c}_\mathrm{q}}(\bm{b}_n-\Bar{\bm{b}})^2},
    H_{cls}^\mathrm{q}=-\sum_{\mathrm{n}=1}^{\mathrm{c}_\mathrm{q}}\Bar{s}_n*\log{\Bar{s}_n}.
    \label{supp_eq:sigentropy}
\end{equation}
Here, $\mathrm{c}_\mathrm{q}$ is the number of members in each list, $q$ is the index of each list, and $\Bar{.}$ denotes the mean of the underlying variable. This technique is referred to as \textbf{MCDO-NMS} which is divided into \textbf{MCDO-NMS-localization}, referred to as $\sigma_b^\mathrm{q}$, and \textbf{MCDO-NMS-classification}, referred to as $H_{cls}^\mathrm{q}$ in \cref{supp_eq:sigentropy}. Inspired by \cite{oksuz2023towards}, which suggests averaging individual uncertainties as one possible aggregation solution, we take a weighted average of classification entropy. Similarly, we sum the square residuals of localization parameters, take a weighted average, and calculate the square root at the end.

\subsection{Benchmark Scenarios}

\subsubsection{Normalization}
\begin{figure}
    \centering
    \includegraphics[width=\linewidth, height=0.9\linewidth]{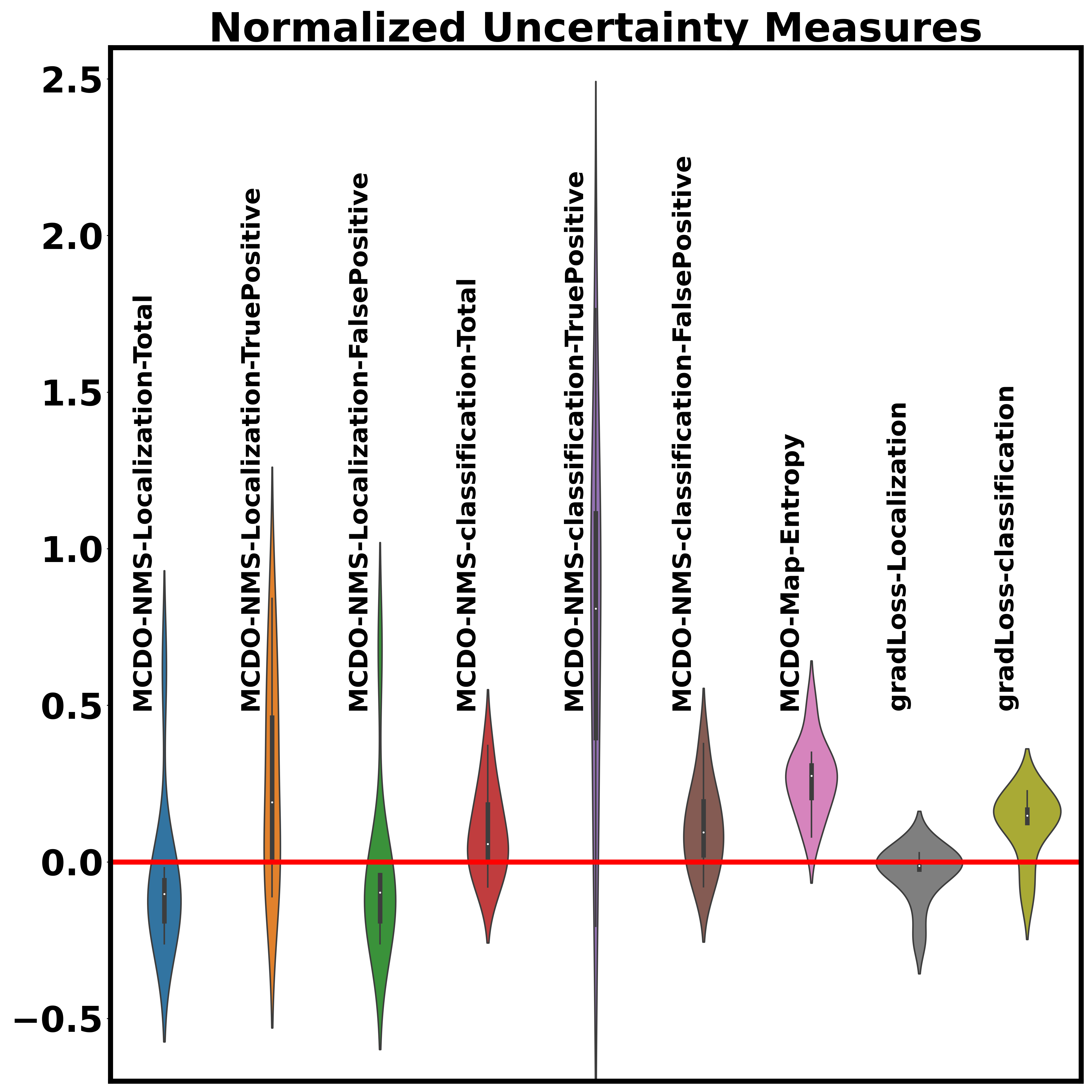}
    \caption{Violin plot of uncertainty metrics: The metrics are normalized to the source domain's uncertainty. MCDO-map always shows positive values, increasing along with DSs, while most of the metrics show negative values.}
    \label{supp_fig:uncmeas}
\end{figure}

Normalization has been done for each metric by subtracting a reference value, which is the value of the metric for the source domain, and then dividing by the same reference value. The normalization is mathematically expressed by
\begin{equation}
    M_{\text{norm}}^{\mathrm{i}} = \frac{M^\mathrm{i} - M_\mathrm{r}}{M_\mathrm{r}}.
\end{equation}
For a set of values of a metric $\mathcal{M} = \{M^1, M^2, \ldots, M^\mathrm{n}\}$ and corresponding reference value $M_\mathrm{r}$, the normalized set is $\mathcal{M}_{\text{norm}} = \{M_{\text{norm}}^{1}, M_{\text{norm}}^{2}, \ldots, M_{\text{norm}}^{\mathrm{n}}\}$.
The subtraction of the reference value $M_\mathrm{r}$ ensures that the data is centered around zero, and the subsequent division by $M_\mathrm{r}$ scales the data, making it comparable or suitable for further analysis. \cref{supp_fig:uncmeas} is illustrated using normalized values of different metrics. All metrics are normalized to their values for the source domain. Positive values indicate increases.

\subsubsection{Violin box plot}
Violin box plot \cite{hullman2015hypothetical} is a graphical representation that combines aspects of both box plots and kernel density plots. It provides a concise and informative way to visualize the distribution, central tendency, and spread of a variable.
In the violin box plot:
\begin{itemize}
    \item The central box represents the interquartile range (IQR) of the data, with the line inside indicating the median.
    \item The "violin" shape surrounding the box displays the probability density function of the data, providing insights into the distribution's shape.
    \item Wider sections of the violin indicate higher data density, while narrower sections represent lower density.
    \item Outliers, if any, are often displayed as individual points.
\end{itemize}

\subsubsection{Pearson correlation coefficient}
Pearson correlation coefficient \cite{cohen2009pearson}, denoted by $\rho$, is a measure of the linear relationship between two variables $\mathcal{M}^1$ and $\mathcal{M}^2$. It is defined as the ratio of the covariance of $\mathcal{M}^1$ and $\mathcal{M}^2$ to the product of their standard deviations,

\begin{equation}
\rho_{\mathcal{M}^1 \mathcal{M}^2} = \frac{cov(\mathcal{M}^1, \mathcal{M}^2)}{\sigma_{\mathcal{M}^1} \sigma_{\mathcal{M}^2}}.
\end{equation}

The Pearson correlation coefficient ranges from -1 to 1. A value of 1 indicates a perfect positive linear relationship, 0 indicates no linear relationship, and -1 indicates a perfect negative linear relationship. Positive values indicate that as one variable increases, the other variable tends to increase as well. Negative values indicate that as one variable increases, the other variable tends to decrease. This coefficient has been used in Fig. 3 of the main paper to analyze the relationships between different metrics.

\subsubsection{Kullback-Leibler (KL) divergence}
KL divergence\cite{kullback1951kullback} is a measure of how one probability distribution diverges from a second, expected probability distribution. In the DrIFT benchmark, it serves as a metric to quantify the distance between feature map distributions of different domains with the source domain, which is ground-synthetic-winter-normal-sky. The KL divergence is defined
\begin{equation}
    \begin{gathered}
        D_{\text{KL}}(\bm{FMD}_{\text{target}} \| \bm{FMD}_{\text{source}}) = \\
        \sum_{\mathrm{i}}^{\mathrm{N}} \bm{FMD}_{\text{target}}^{\mathrm{i}} \log\left(\frac{\bm{FMD}_{\text{target}}^{\mathrm{i}}}{\bm{FMD}_{\text{source}}^{\mathrm{i}}}\right),
    \end{gathered}
\end{equation}
in which N is the cardinality of the feature map distributions, $ |\bm{FMD}_{\text{target}}| $. We assume the two distributions have the same size, $ |\bm{FMD}_{\text{target}}| = |\bm{FMD}_{\text{source}}| $. $\bm{FMD}_{\text{target}}$ is the feature map distribution of each domain that is taken as the target domain, and $\bm{FMD}_{\text{source}}$ is the source domain's feature map distribution. i is the index of existing elements in each domain's feature map distribution.
\subsection{Experiments and Results}
\label{supp_subsec:exp_res}
The ground-synthetic-winter-normal-sky is taken as the source domain all over the paper and supplementary material unless we specify other domains.

\subsubsection{Implementation Details}
For the object detector in this work, the Faster R-CNN \cite{ren2015faster} architecture with a VGG16 \cite{simonyan2014very} in the mmdetection platform \cite{MMDetection_Contributors_OpenMMLab_Detection_Toolbox_2018} has been utilized. For generalization and MC-dropout uncertainty evaluation implementation, the dropout has been activated within the VGG. The experiments were run on a Desktop with a Geforce RTX 3090 and a High-performance computing cluster providing 4 x NVidia A100 (40 GB memory). For the vanilla network training that was started from scratch, we used a stochastic gradient descent optimizer for 73 epochs, for which the learning rate was 0.24 for a batch size of 6 on each GPU. For adaptation training, the vanilla network is used as the pre-trained weights. The learning rate has been decreased to $10^{-5}$, and the discriminator, which is a simple convolutional neural network, has been trained by an Adam optimizer with a learning rate of $10^{-6}$. The codes and details will be available on an online platform.

The objective of \cref{supp_tab:unccomp} and \cref{supp_tab:unccomp2} is to compare our uncertainty estimation method, \textbf{MCDO-map}, with various uncertainty estimation metrics mentioned in the main paper. In \cref{supp_tab:unccomp}, the source domain was ground-synthetic-winter-normal-sky, while ground-real-winter-normal-sky served as the source domain in \cref{supp_tab:unccomp2}. To provide a comprehensive explanation, we utilized \cref{supp_fig:corr} to discover a meaningful relation between different uncertainty evaluation metrics, AP, D-ECE, and KL divergence metric (which measures the distance between feature map distributions of different domains relative to the source domain) using the Pearson correlation coefficient. The findings in \cref{supp_fig:corr} could be summarized as follows:

\begin{itemize}
    \item MCDO-map exhibits the highest positive correlation (0.81) with KL divergence, indicating its superior capability to capture DSs. A greater level of shift, reflected by increased distance or KL divergence, correlates with higher values of MCDO-map.
    \item As an uncertainty evaluation metric, a negative correlation with AP is expected, implying that higher AP values correspond to lower uncertainty levels. In this context, MCDO-NMS-Loc-Total, MCDO-NMS-Loc-FP, and MCDO-map yield the best results.
    \item Positive correlations between D-ECE and most uncertainty evaluation metrics suggest that increased uncertainty tends to coincide with calibration errors.
    \item A positive correlation between D-ECE and AP indicates that even with higher AP values, the model may exhibit over or under-confidence, compromising its reliability.
    \item Positive correlations between D-ECE and most uncertainty evaluation metrics, such as 0.36 for MCDO-map, suggest that higher levels of uncertainty are associated with calibration errors.
\end{itemize}

Consequently, MCDO-map emerges as a wise choice for our UDA algorithm to capture DSs effectively.

\begin{table*}
    \centering
    {\small{
    \begin{tabular}{ccccc|lllllll|l|ll}
        \toprule
        \multicolumn{5}{c|}{Validation Domain} & \multicolumn{7}{c|}{MCDO-NMS$\times10^{-3}$} & MCDO-Map & \multicolumn{2}{c}{grad-loss$\times10^{-3}$} \\
        \multicolumn{5}{c|}{} & \multicolumn{3}{c}{Localization} & & \multicolumn{3}{c|}{Classification} & $\times10^{-4}$ & Loc. & Cls. \\
        \cmidrule{1-8}
        \cmidrule{10-15}
        View & Source & Season & Weather & BG & Total & TP & FP && Total & TP & FP &  &  & \\
        \midrule
        ground & synthetic & winter & normal & - & 107 & 83 & 107 &&  457 & 354 & 459 & 3582 & 433 & 610 \\
        ground & synthetic & winter & normal & sky & 107 & 63 & 107 &&  436 & 141 & 439 & - & 445 & 564 \\
        ground & synthetic & winter & normal & \cellcolor{lightgray}tree & 108 & 93 & 108 &&  512 & 523 & 512 & - & 451 & 680 \\
        ground & synthetic & winter & normal & \cellcolor{lightgray}ground & 106 & 104 & 107 &&  440 & 476 & 422 & - & 397 & 670 \\
        \midrule
        \midrule
        ground & \cellcolor{lightgray}real & winter & normal & - & 183 & 56 & 186 && 492 & 313 & 496 & 5355 & 477 & 708 \\
        ground & \cellcolor{lightgray}real & winter & normal & sky & 173 & 56 & 180 && 495 & 298 & 508 & - & 434 & 692 \\
        ground & \cellcolor{lightgray}real & winter & normal & \cellcolor{lightgray}tree & 162 & 100 & 163 && 442 & 689 & 432 & - & 451 & 680 \\
        ground & \cellcolor{lightgray}real & winter & normal & \cellcolor{lightgray}ground & 189 & 59 & 189 && 490 & 578 & 490 & - & 502 & 717 \\
        \midrule
        \midrule
        ground & synthetic & adverse & \cellcolor{lightgray}rainy & sky &  98 & 67 & 99 && 401 & 112 & 404 & 3866 & 390 & 634 \\
        ground & synthetic & adverse & \cellcolor{lightgray}snowy & sky & 79 & 58 & 79 && 454 & 183 & 456 & 4686 & 458 & 658 \\
        ground & synthetic & adverse & \cellcolor{lightgray}foggy & sky & 85 & 83 & 85 && 449 & 241 & 484 & 4454 & 335 & 642 \\
        \midrule
        \midrule
        \cellcolor{lightgray}aerial & synthetic & winter & normal & - & 100 & 78 & 101 && 436 & 335 & 438 & 4287 & 373 & 601 \\
        \cellcolor{lightgray}aerial & synthetic & winter & normal & sky &  101 & 62 & 103 && 407 & 166 & 417 & - & 444 & 500 \\
        \cellcolor{lightgray}aerial & synthetic & winter & normal & \cellcolor{lightgray}tree & 104 & 74 & 104 &&  499 & 463 & 499 & - & 415 & 659 \\
        \cellcolor{lightgray}aerial & synthetic & winter & normal & \cellcolor{lightgray}ground & 98 & 90 & 99 &&  426 & 427 & 426 & - & 341 & 628 \\
        \midrule
        \cellcolor{lightgray}aerial & synthetic & \cellcolor{lightgray}fall & normal & - & 95 & 99 & 95 && 509 & 351 & 515 & 4680 & 424 & 650 \\
        \cellcolor{lightgray}aerial & synthetic & \cellcolor{lightgray}fall & normal & sky & 92 & 105 & 92 && 523 & 312 & 529 & - & 435 & 653 \\
        \cellcolor{lightgray}aerial & synthetic & \cellcolor{lightgray}fall & normal & \cellcolor{lightgray}tree & 100 & 59 & 101 && 492 & 386 & 495 & - & 391 & 658 \\
        \cellcolor{lightgray}aerial & synthetic & \cellcolor{lightgray}fall & normal & \cellcolor{lightgray}ground & 102 & 107 & 102 && 456 & 453 & 456 & - & 427 & 611 \\
        \midrule
        \cellcolor{lightgray}aerial & synthetic & \cellcolor{lightgray}summer & normal & - & 95 & 92 & 95 && 525 & 299 & 532 & 4677 & 424 & 657 \\
        \cellcolor{lightgray}aerial & synthetic & \cellcolor{lightgray}summer & normal & sky & 94 & 92 & 94 && 538 & 269 & 543 & - & 434 & 658 \\
        \cellcolor{lightgray}aerial & synthetic & \cellcolor{lightgray}summer & normal & \cellcolor{lightgray}tree & 108 & 104 & 112 && 478 & 222 & 505 & - & 364 & 642 \\
        \cellcolor{lightgray}aerial & synthetic & \cellcolor{lightgray}summer & normal & \cellcolor{lightgray}ground & 93 & 116 & 91 && 465 & 464 & 465 & - & 410 & 663 \\
        \midrule
        \midrule
        \cellcolor{lightgray}aerial & synthetic & adverse & \cellcolor{lightgray}rainy & sky & 100 & 092 & 101 &&  440 & 238 & 446 & 4084 & 468 & 554 \\
        \cellcolor{lightgray}aerial & synthetic & adverse & \cellcolor{lightgray}snowy & sky & 85 & 67 & 85 &&  598 & 298 & 605 & 4835 & 445 & 690 \\
        \cellcolor{lightgray}aerial & synthetic & adverse & \cellcolor{lightgray}foggy & sky & 105 & 116 & 103 &&  468 & 390 & 477 & 4394 & 445 & 635 \\
        \bottomrule
    \end{tabular}
    }}
    \caption{Comparison of our uncertainty estimation metric, \textbf{MCDO-map}, with other methods for the Faster R-CNN trained on the source domain. Each row shows the validation domain. In each row, three different methods have evaluated the uncertainty level. MCDO-NMS reported separately for TP and FP detections. MCDO-map works better than other methods in terms of capturing DSs effectively. Loc.: localization, Cls.: Classification}
    \label{supp_tab:unccomp}
\end{table*}
\begin{figure*}
    \centering
    \begin{subfigure}{0.49\linewidth}
        \includegraphics[width=\linewidth]{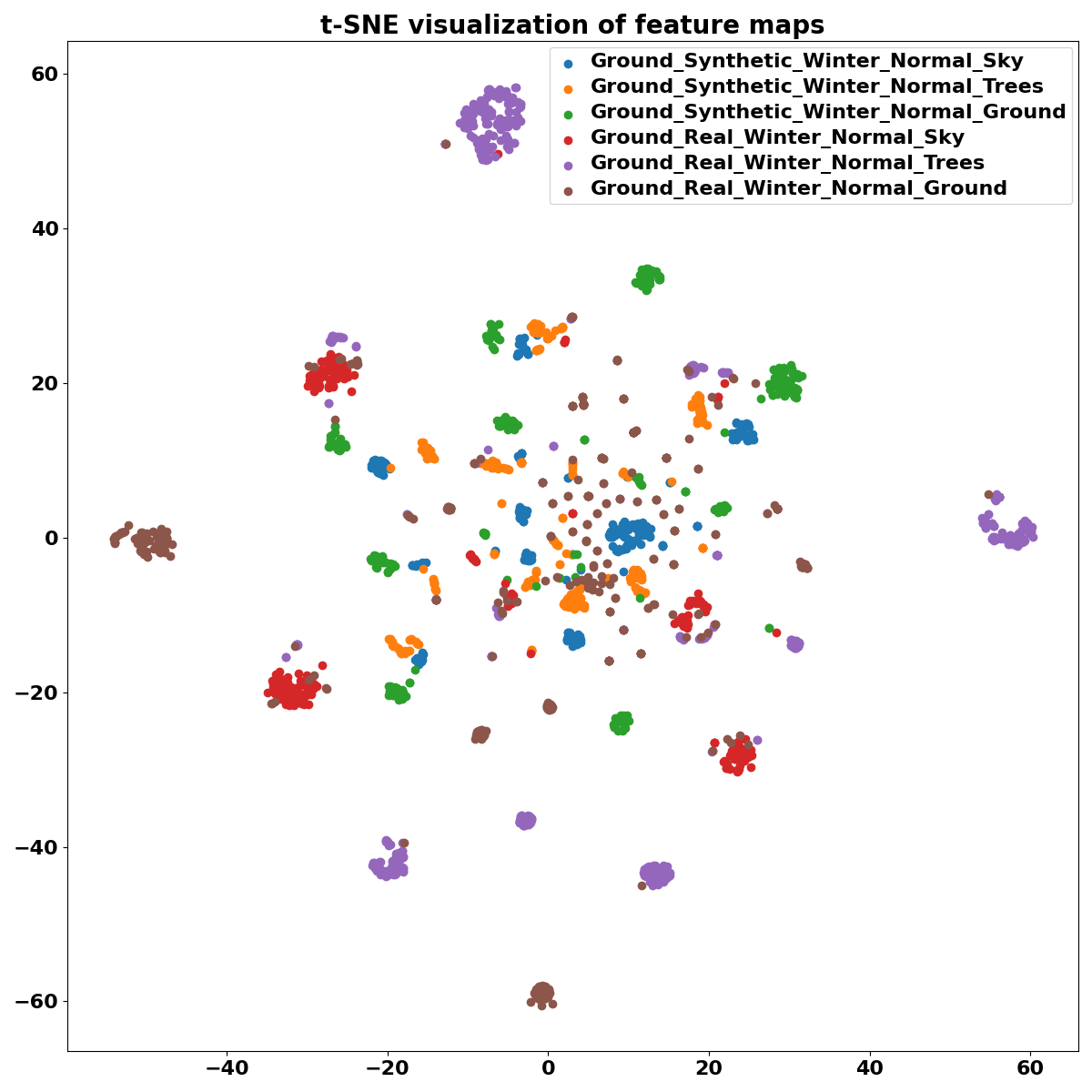}
        \caption{}
        \label{supp_fig:featmapdist1}
    \end{subfigure}
    \begin{subfigure}{0.49\linewidth}
        \includegraphics[width=\linewidth]{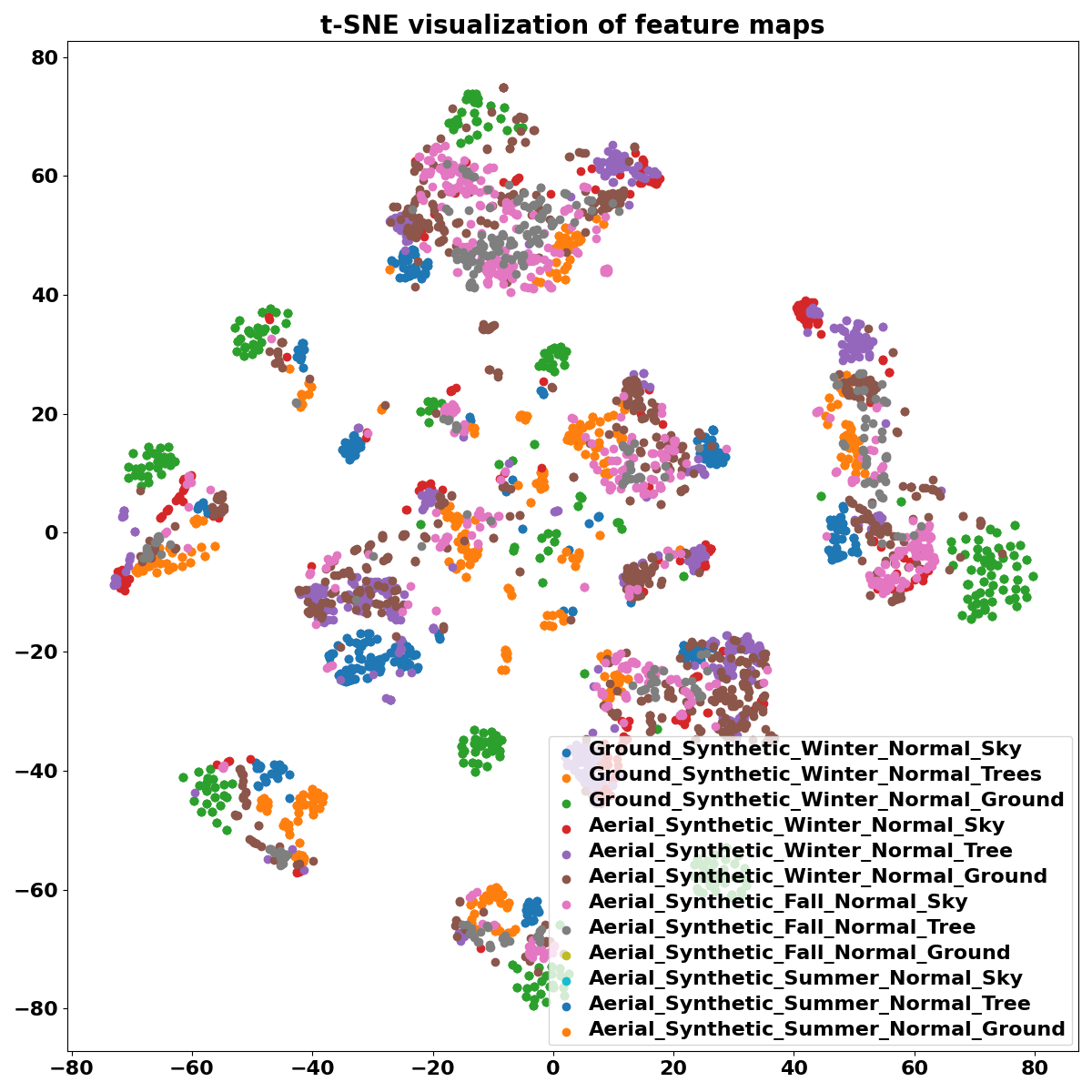}
        \caption{}
        \label{supp_fig:featmapdist2}
    \end{subfigure}
    \hfill
    \begin{subfigure}{0.49\linewidth}
        \includegraphics[width=\linewidth]{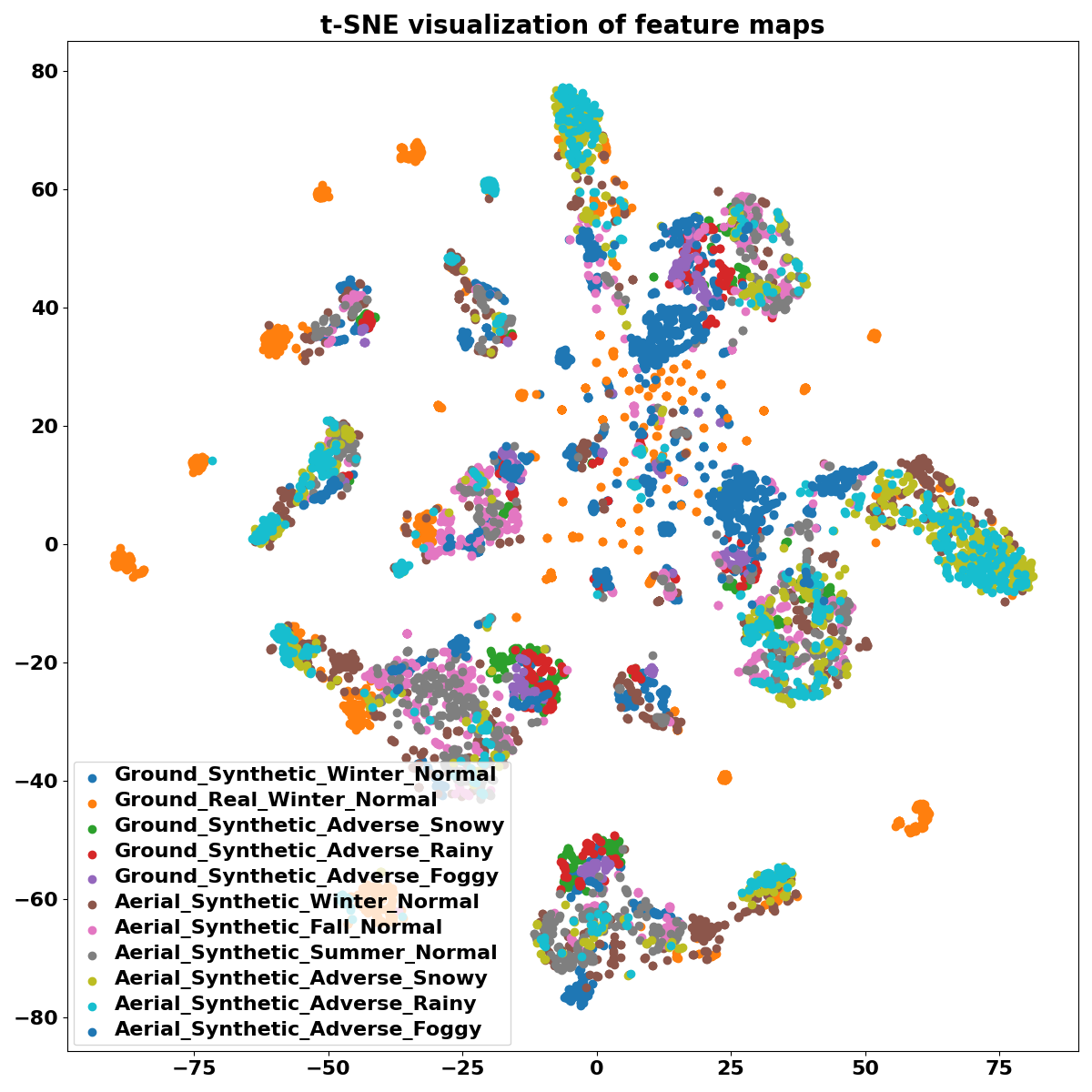}
        \caption{}
        \label{supp_fig:featmapdist3}
    \end{subfigure}
    \caption{2D representation of Faster R-CNN's, trained on the source domain, last layer feature maps distribution for different domains by utilizing t-SNE. a) For the target domains, synthetic data is changed to real data with sky, tree, and ground backgrounds. b) The view and season have been changed to aerial view, and fall and summer, respectively, as well as different backgrounds. c) All defined domains are taken into account without separating the different backgrounds.}
    \label{supp_fig:featmapdist}
\end{figure*}
\begin{figure*}
    \centering
    \includegraphics[width=\linewidth]{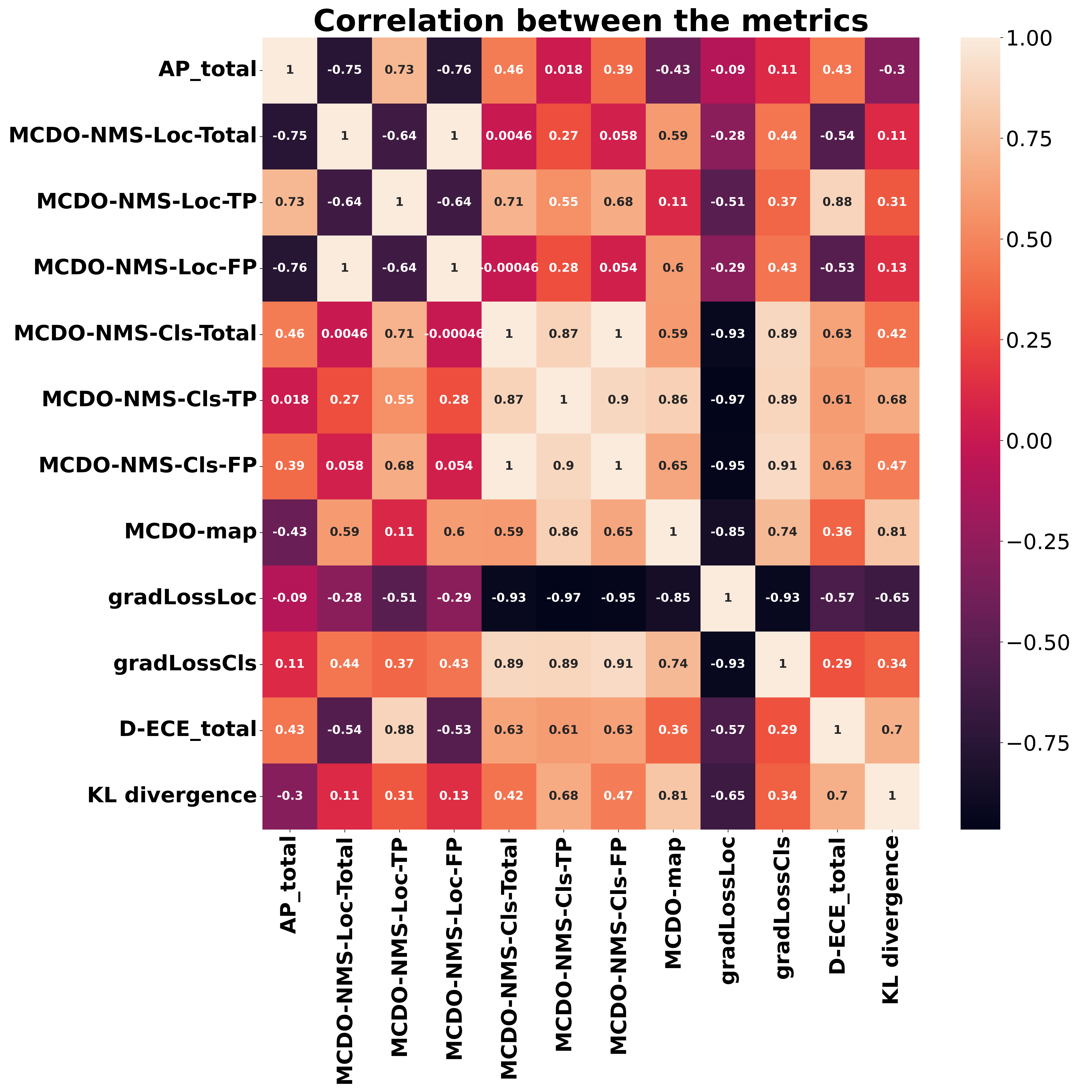}
    \caption{Correlation heatmap of metrics in DrIFT: MCDO-map exhibits the highest positive correlation (0.81) with KL divergence, indicating its superior capability to capture DSs. MCDO-NMS-Loc-Total, MCDO-NMS-Loc-FP, and MCDO-map yield the top three negative correlations with AP. MCDO-map emerges as the best metric in terms of capturing DSs effectively.}
    \label{supp_fig:corr}
\end{figure*}

\begin{table*}
    \centering
    {\small{
    \begin{tabular}{ccccc|lllllll|l|ll}
        \toprule
        \multicolumn{5}{c|}{Validation Domain} & \multicolumn{7}{c|}{MCDO-NMS$\times10^{-3}$} & MCDO-Map & \multicolumn{2}{c}{grad-loss$\times10^{-3}$} \\
        \multicolumn{5}{c|}{} & \multicolumn{3}{c}{Localization} & & \multicolumn{3}{c|}{Classification} & $\times10^{-4}$ & Loc. & Cls. \\
        \cmidrule{1-8}
        \cmidrule{10-15}
        View & Source & Season & Weather & BG & Total & TP & FP && Total & TP & FP &  &  & \\
        \midrule
        ground & real & winter & normal & - & 84 & 31 & 88 && 405 & 146 & 425 & 5355 & 299 & 621 \\
        ground & real & winter & normal & sky & 173 & 56 & 180 && 495 & 298 & 508 & - & 434 & 692 \\
        ground & real & winter & normal & \cellcolor{lightgray}tree & 86 & 34 & 86 && 464 & 489 & 464 & - & 358 & 675 \\
        ground & real & winter & normal & \cellcolor{lightgray}ground & 71 & 79 & 71 && 377 & 388 & 377 & - & 454 & 564 \\
        \bottomrule
    \end{tabular}
    }}
    \caption{Comparison of our uncertainty estimation metric, \textbf{MCDO-map}, with other methods for the Faster R-CNN trained on ground-real-winter-normal-sky domain. Each row shows the validation domain. In each row, three different methods have evaluated the uncertainty level. MCDO-NMS reported separately for TP and FP detections. MCDO-map works better rather than other methods in terms of capturing DSs effectively. Loc.: localization, Cls.: Classification}
    \label{supp_tab:unccomp2}
\end{table*}
\begin{figure*}
  \centering
  \begin{subfigure}{\linewidth}
    \includegraphics[width=\linewidth]{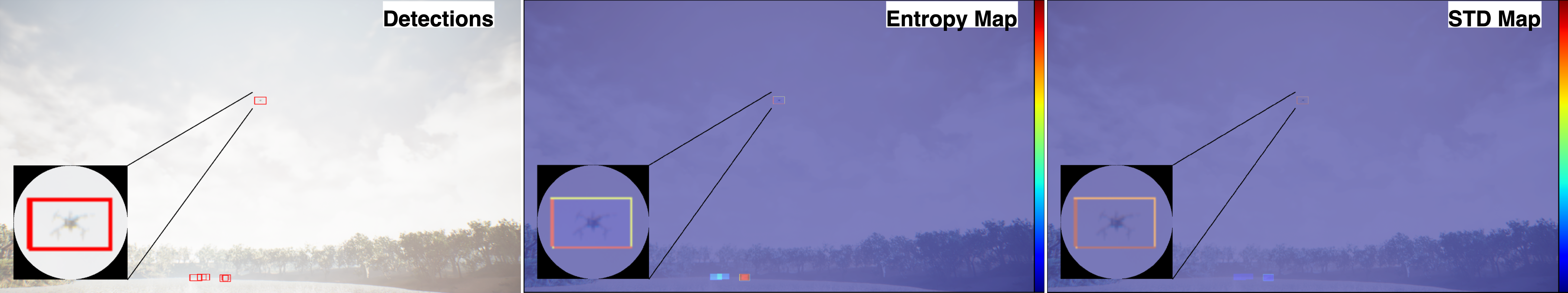}
    \caption{Target drone with sky background}
    \label{supp_fig:res1}
  \end{subfigure}
  \begin{subfigure}{\linewidth}
    \includegraphics[width=\linewidth]{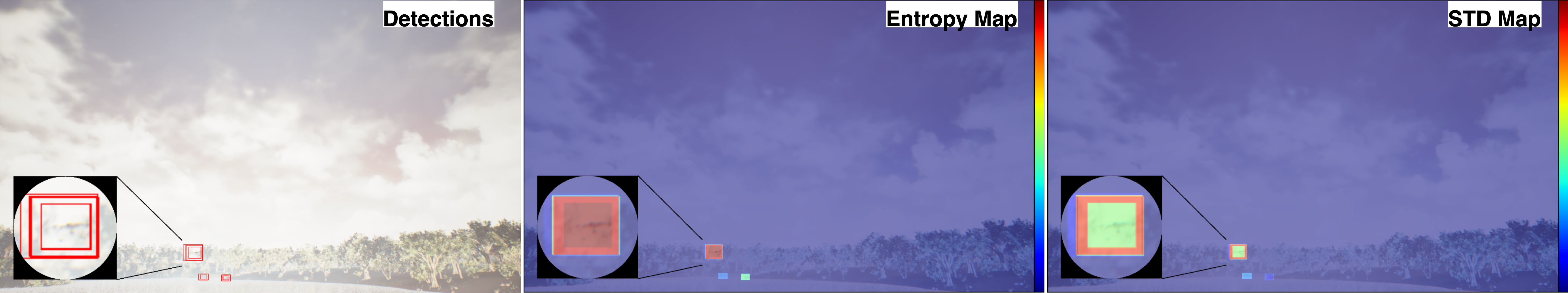}
    \caption{Target drone with tree background}
    \label{supp_fig:res2}
  \end{subfigure}
  \begin{subfigure}{\linewidth}
    \includegraphics[width=\linewidth]{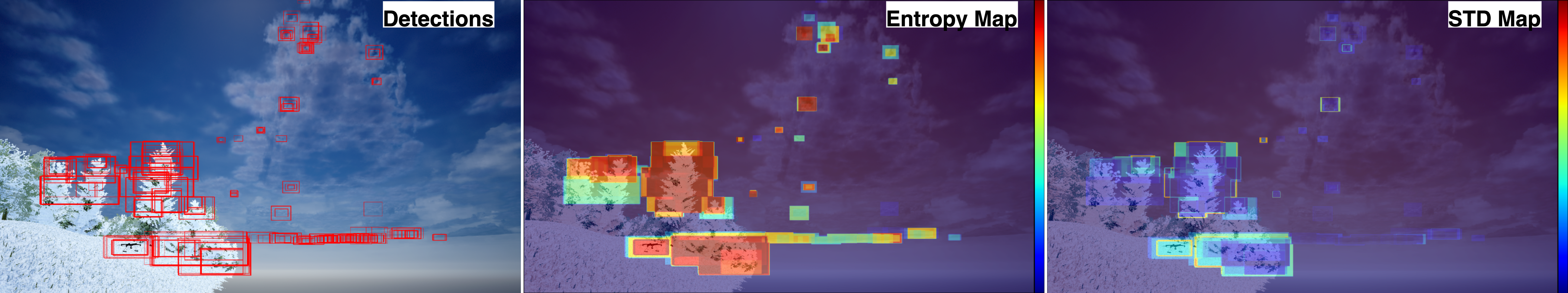}
    \caption{Target drone with ground background}
    \label{supp_fig:res3}
  \end{subfigure}
  \hfill
  \caption{Visual results of the trained faster R-CNN on the Ground-Synthetic-Winter-Normal-Sky domain: In (a), the drone with a sky background exhibits a low level of uncertainty, as indicated by the bounding box being entirely blue in the entropy map, with non-zero std values only at the edge. In (b), the drone with a tree background demonstrates a higher level of uncertainty, with red areas in the entropy map and non-zero standard deviation values within the bounding box. In (c), the drone with a ground background is detected with the highest level of uncertainty in this figure, with an intense red coloration within the box in the entropy map and high std.}
  \label{supp_fig:res}
\end{figure*}
To enhance the understanding of our results, we present three examples of the outputs generated by the trained Faster R-CNN model on the ground-synthetic-winter-normal-sky domain, depicted in \cref{supp_fig:res}. In \cref{supp_fig:res1}, the drone with a sky background exhibits low uncertainty, as indicated by the blue bounding box on the entropy map. We observe non-zero std values only at the edge of the bounding box in the std map (inside blue, red at the edge). However, a few false detections occur during the MCDO iterations, resulting in non-zero values in both maps around the intersection of the tree and ground. Moving to \cref{supp_fig:res2}, the drone with a tree background demonstrates higher uncertainty. The bounding box exhibits some red areas in the entropy map, accompanied by non-zero standard deviation values inside the bounding box. Once again, false detections contribute to non-zero values in the maps. Finally, in \cref{supp_fig:res3}, the drone with a ground background is detected with the highest level of uncertainty among these cases, corresponding to way too red color for the bounding box in the entropy map and nonzero values within the bounding box in the std map. However, a significant number of false detections around trees contribute to a considerable level of uncertainty in the maps, reflecting the low AP for trees and, consequently, higher uncertainty in this domain. Detailed AP and uncertainty values for trees are provided in Tab. 2 of the main paper.

\end{document}